\documentclass[10pt,twocolumn,letterpaper]{article}

\usepackage{cvpr}              

\usepackage{graphicx}
\usepackage{amsmath}
\usepackage{amssymb}
\usepackage{booktabs}
\usepackage{multirow}
\usepackage{pifont}
\usepackage{makecell}
\usepackage[table]{xcolor}

\usepackage[pagebackref,breaklinks,colorlinks]{hyperref}
\usepackage[accsupp]{axessibility}
\begin{document}

\title{Visible-Thermal UAV Tracking: A Large-Scale Benchmark and New Baseline}

\author{Pengyu Zhang$^1$, Jie Zhao$^1$, Dong Wang$^{1\dag}$, Huchuan Lu$^{1,2}$, Xiang Ruan$^3$\\
$^1$School of Information and Communication Engineering, Dalian University of Technology, China\\
$^2$Peng Cheng Laboratory $^3$Tiwaki Co.Ltd.\\
{\tt\small {\{pyzhang, zj982853200\}@mail.dlut.edu.cn}, \{wdice, lhchuan\}@dlut.edu.cn, ruanxiang@tiwaki.com}
}
\maketitle

\begin{abstract}
With\let\thefootnote\relax\footnotetext{$\dag$ Corresponding author: Dr. Dong Wang, wdice@dlut.edu.cn} the popularity of multi-modal sensors, visible-thermal~(RGB-T) object tracking is to achieve robust performance and wider application scenarios with the guidance of objects' temperature information. However, the lack of paired training samples is the main bottleneck for unlocking the power of RGB-T tracking. Since it is laborious to collect  high-quality RGB-T sequences, recent benchmarks only provide test sequences. In this paper, we construct a large-scale benchmark with high diversity for visible-thermal UAV tracking~(VTUAV), including 500 sequences with 1.7 million high-resolution~($1920*1080$ pixels) frame pairs. In addition, comprehensive applications~(short-term tracking, long-term tracking and segmentation mask prediction) with diverse categories and scenes are considered for exhaustive evaluation.  Moreover, we provide a coarse-to-fine attribute annotation, where frame-level attributes are provided to exploit the potential of challenge-specific trackers. In addition, we design a new RGB-T baseline, named Hierarchical Multi-modal Fusion Tracker~(HMFT), which fuses RGB-T data in various levels. Numerous experiments on several datasets are conducted to reveal the effectiveness of HMFT and the complement of different fusion types. The project is available at \href{https://zhang-pengyu.github.io/DUT-VTUAV/}{here}.
\end{abstract}

\section{Introduction}\label{sec:intro}
\begin{figure}[t]
    \centering
    \includegraphics[width=0.98\linewidth]{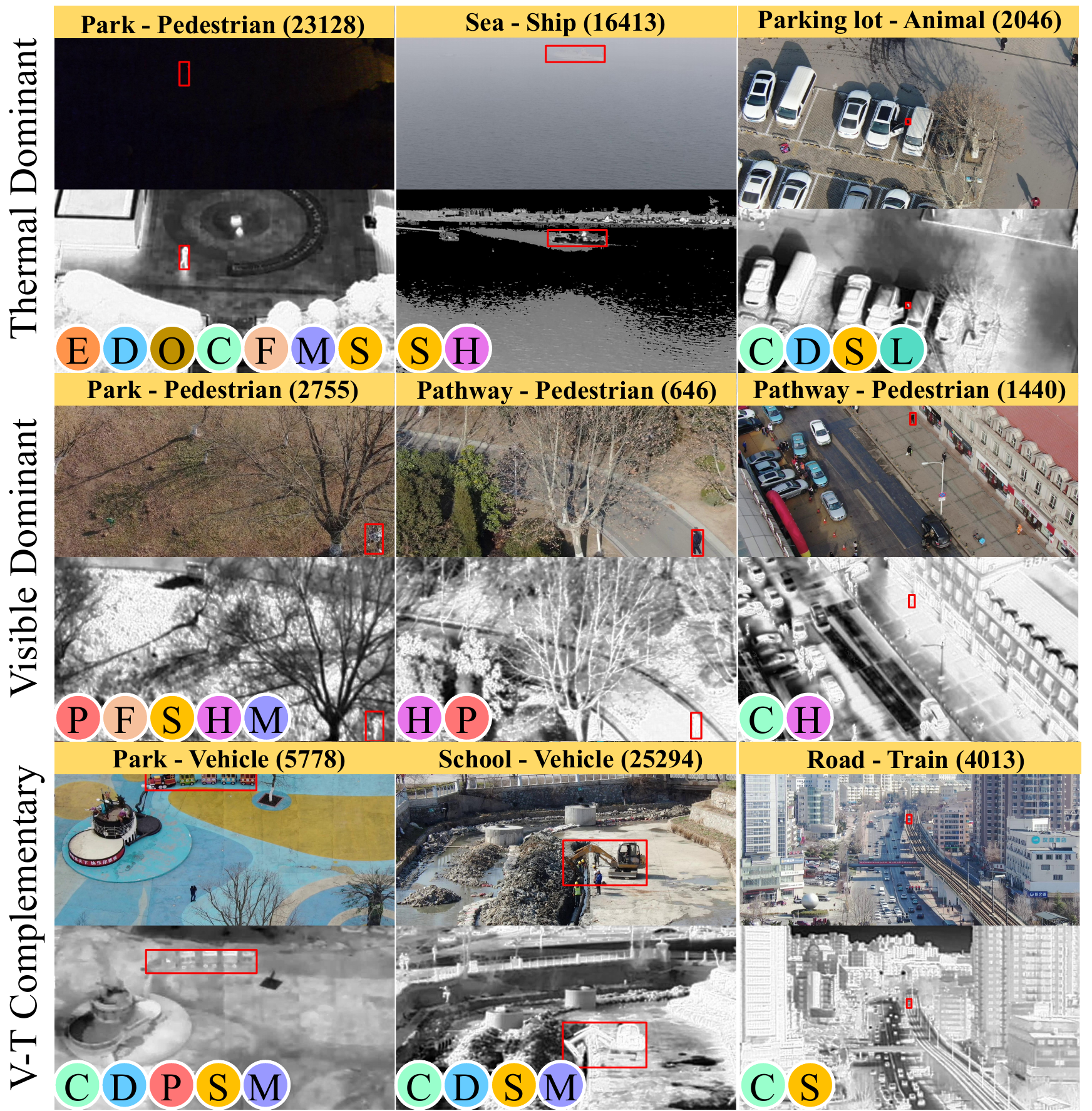}
    \caption{Sample frames in our dataset. \emph{Scenes - super class (sequence length)} are shown on the top. Sequence-level attributes are shown at bottom, including
    camera movement~(C), deformation~(D), 
    extreme illumination~(E), 
    partial occlusion~(P),
    full occlusion~(F), 
    scale variation~(S), 
    thermal clustering~(H), 
    fast moving~(M), 
    out-of-view~(O), and
    low resolution~(L).
    }

    \label{fig:sample}
\end{figure}

Given the initial position of a model-agnostic target, visual object tracking is to capture the target in the subsequent frames~\cite{Tracking-Book}, where the target may suffer out-of-view, occlusion, illumination variation, and motion blur. Previous algorithms solve those challenges within visible modality, providing limited information when the target is in dark, rainy, foggy and other extreme conditions (the first row in Fig.~\ref{fig:sample}). By contrast, thermal image, as a complementary cue, is insensitive to illumination variation, while it cannot distinguish the target when the target and background are in similar temperature (the second row in Fig.~\ref{fig:sample}). To this end, with the portability and low-price of multi-modality sensors, tracking with visible-thermal~(RGB-T) data enlarges the application scope by providing complementary information, which has attached more attentions~\cite{Zhang_Arxiv20_MM_tracking_survey,Li_NEU18_FTSNet}.
Li \emph{et al.}~\cite{Li_TIP16_GTOT} release a gray-scale RGB-T dataset with 50 videos. Later, RGBT210~\cite{Li_MM17_RGBT210} and RGBT234~\cite{Li_PR19_RGBT234} are proposed, containing 210 and 234 test videos. In 2019~\cite{Kristan_ICCVW19_VOT19} and 2020~\cite{Kristan_ECCVW20_VOT20}, VOT committee holds the VOT-RGBT tracking subchallenge, which selects 60 sequences from RGBT234 to evaluate the accuracy and robustness of competitors. Furthermore, various algorithms are proposed by considering performance and time cost. Li \emph{et al.}~\cite{Li_ICCVW19_MANet} propose a Multi-Adaptor network to learn modality-shared and modality-specific representations. Zhang \emph{et al.}~\cite{Zhang_IJCV21_ADRNet} design a real-time RGB-T tracker that exploits the effectiveness of attribute annotation. Zhang \emph{et al.}~\cite{Zhang_ICCVW19_DIMP-RGBT} extend DiMP~\cite{Bhat_ICCV19_DIMP} to RGB-T tracking, obtaining the best ranking in VOT2019-RGBT.

However, the lack of training data becomes the main bottleneck for RGB-T tracking. Existing datasets~(GTOT, RGBT210, RGBT234, and VOT-RGBT) contain 284 unique short-term sequences overall. Trackers have to be trained on another test set~\cite{Li_ICCVW19_MANet,Zhang_IJCV21_ADRNet} or synthetic data generated from visible modality~\cite{Zhang_ICCVW19_DIMP-RGBT,Zhang_TCSVT_SiamCDA}, which suffer limited generalization ability and training gap. Moreover, test sequences are captured with monitoring devices, thereby leading to limited viewpoint, frame length, and imaging quality. 
To fully exploit the potential of the RGB-T tracker, this paper presents a large-scale RGB-T tracking dataset with high diversity. The main contributions are listed as follows:

\begin{itemize}
    \item We construct a  large-scale  benchmark  with  high  diversity  for  visible-thermal  UAV  tracking~(VTUAV). To our best knowledge, VTUAV is the largest multi-modal tracking dataset with the highest resolution. Moreover, we take short-term, long-term tracking and segmentation mask prediction into consideration to achieve a comprehensive evaluation with wider applications. We also provide an exquisite attribute annotation in frame and sequence levels, which can meet the requirement of training a challenge-specific tracker.

    \item We propose a new baseline for RGB-T tracking, namely HMFT, which unifies various multi-modal fusion strategies (including image fusion, feature fusion and decision fusion) into a hierarchical fusion framework. We implement corresponding versions for short-term and long-term tracking. Furthermore, we provide an in-depth analysis on various fusion types to develop RGB-T trackers. Exhaustive experiments on GTOT, RGBT210, RGBT234 and VTUAV conclude the complement of various fusion types.
\end{itemize}

\section{Related Work}
\noindent\textbf{RGB-T tracking benchmarks.}
The first dataset used for RGB-T tracking is OTCBVS~\cite{Davis_CVIU07_OTCBVS}, which contains 6 sequences with 7200 frames. In 2012, LITIV~\cite{Torabi_CVIU12_LITIV} was proposed with 9 video clips and 6300 image pairs. These two datasets are out-of-date since they are not particularly designed for RGB-T tracking and with limited data. In 2016, Li \emph{et al.}~\cite{Li_TIP16_GTOT} propose a grayscale-thermal tracking dataset, namely GTOT, which contains 7800 frames. GTOT contains various challenging scenes to evaluate the trackers' robusteness in extreme conditions. The RGBT210~\cite{Li_MM17_RGBT210} dataset is released, with 210 videos and  more than 104K frames. Later, RGBT234~\cite{Li_PR19_RGBT234}, an extended version of RGBT210, enlarges the number of sequence to 234 and provides a modality-independent annotation, which can be used to learn individual models separately. In 2019, VOT committee selects 60 sequences and construct a new dataset VOT-RGBT~\cite{Kristan_ICCVW19_VOT19}, which utilizes Expected Average Overlap~(EAO) to evaluate the accuracy and robustness of trackers. The LSS dataset~\cite{Zhang_TCSVT_SiamCDA} is a newly-built synthetic dataset, where either visible or thermal images are generated from another modality using image translation or video colorization methods. Recently, LasHeR~\cite{Li_Arxiv21_LasHeR} contains 1224 short-term videos and 730K frames with multiple scenes and viewpoints. In this paper, we propose a unified large-scale RGB-T tracking dataset with high-quality training pairs. Compared with the most recent dataset~(LasHeR), three main differences can be summarized. First, we have higher quality images and a wider distribution in frame length. Second, LasHeR focuses on short-term evaluation while our dataset measures trackers' performance from three mainstream perspectives including tracking accuracy, target redetection, and pixel-level estimation. Third, detailed 
frame-level attribute annotations are provided, which can meet the requirement of challenge-aware trackers~\cite{Zhang_IJCV21_ADRNet,Li_ECCV20_CAT}.

\noindent\textbf{RGB-T tracking algorithms.}
Recent RGB-T trackers focus on exploiting the correspondence and discriminability of multi-modal information~\cite{Zhang_Sensor20_MaCNet,Zhang_CISP18_MDNet-RGBT,Zhu_arxiv18_FANet,Zhu_MM19_DAPNet}. Several fusion methods are proposed, which can be categorized as image fusion, feature fusion, and decision fusion. 
For image fusion, Peng \emph{et al.}~\cite{Peng_arxiv21_SiamIVFN} utilize a group of layers to learn complementary information by sharing weights for heterogeneous data. Image-fusion-based method can provide a shared representation of multi-modal, while highly depends on image alignment and has not been exploited sufficiently. 
Most trackers aggregate the representation by fusing features~\cite{Wang_arxiv2021_MGFNet}, which can be detailed as two types, i.e., modality interaction and direct fusion.
The former is to refine uni-modal feature guided by another one, and then the features from both modalities are combined, thereby achieving comprehensive representations~\cite{Wang_CVPR20_CMPP,Wang_arxiv2021_MGFNet}.
By contrast, using the multi-modal features as input, the latter combines them first and learns a fused representation by direct concatenation~\cite{Zhang_CISP18_MDNet-RGBT,Zhang_ICCVW19_DIMP-RGBT} or attention technique~\cite{Peng_arxiv21_SiamIVFN}.
Feature fusion has the potential of high flexibility and can be trained with massive unpaired data, which is well-designed to achieve significant promotion.
Decision fusion models each modality independently, and the scores are fused to obtain the final candidate. JMMAC~\cite{Zhang_TIP21_JMMAC} adopts a multi-modal fusion network to ensemble the responses by considering modality-level and pixel-level importance. 
Luo \emph{et al.}~\cite{Luo_IPT19_AWS} utilize independent frameworks to track in RGB-T data, and then the results are combined by adaptive weighting.
Decision fusion avoids the heterogeneity of different modalities and is not sensitive to modality registration.
In this work, we also design a new baseline for RGB-T tracking using hierarchical fusion manner, which derives benefit from all aforementioned three fusion types. Numerous results on three popular RGB-T datasets show that the information fused in various levels can provide comprehensive contributions to obtain a superior result.

\begin{figure*}[t]
    \centering
    \includegraphics[width=0.95\linewidth]{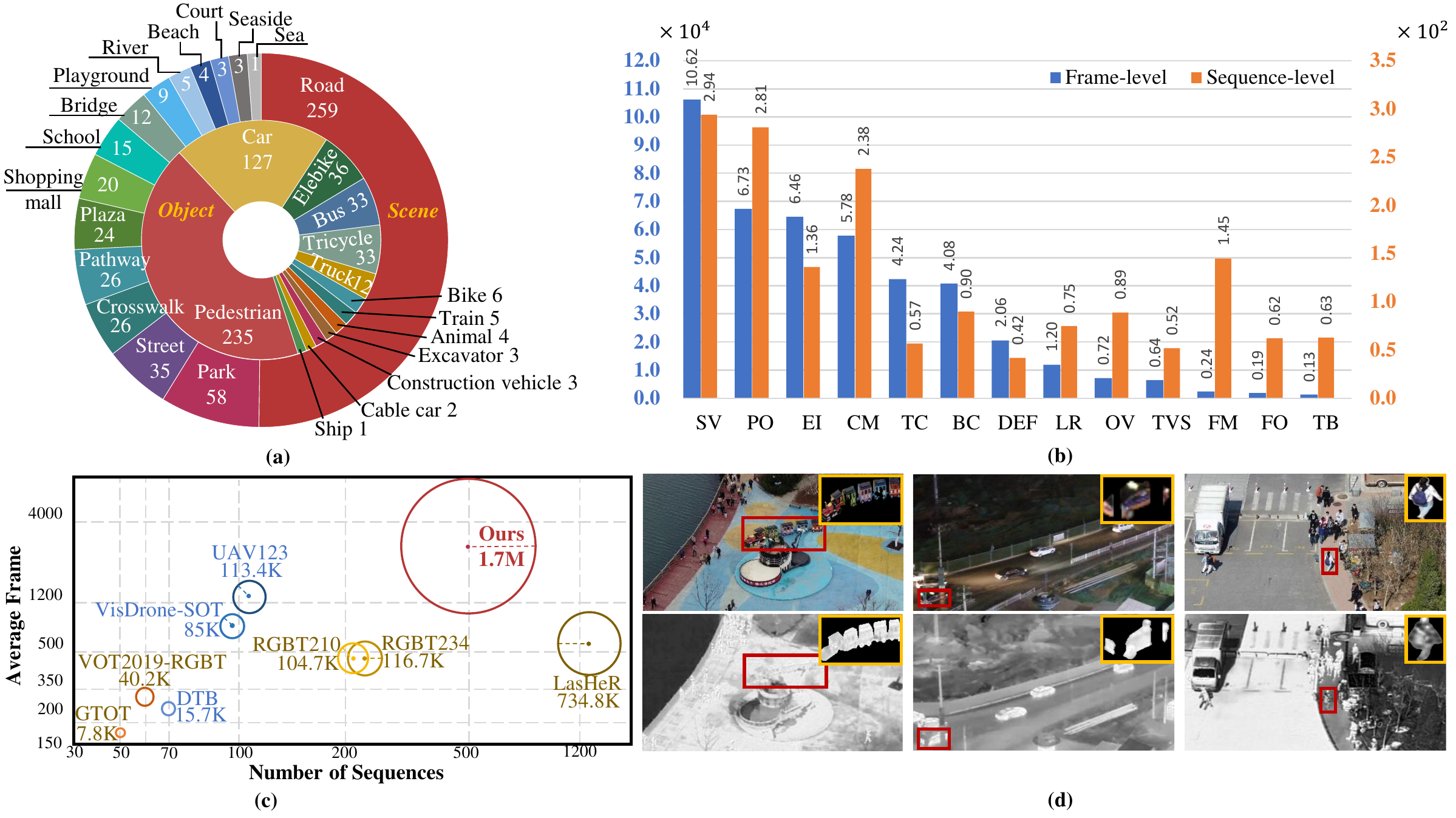}
    \caption{Main features and statistics of the proposed dataset. (a) Distribution of scene~(outer) and object category~(inner). (b) Statistics of frame-level and sequence-level attributes. (c) Comparison among existing datasets and the proposed dataset. The area of each cycle denotes the number of total frame. (d) Precise annotation with bounding box and segmentation mask. Better viewed in color with zoom in.
    }
    \label{datasetDetail}
\end{figure*}

\section{VTUAV Benchmark}

\subsection{Benchmark Features and Statistics}
\begin{itemize}
    \item \emph{Large-scale sequences with high diversity}.
    Recent RGB-T datasets use the multi-sensor surveillance camera with a 2 degree-of-freedom turnable platform. The image quality and flexibility cannot meet the requirements for tracking. Moreover, the target cannot be tracked for a long time with the still camera, leading to a limited frame length.
    To address these issues, our dataset is captured by a professional UAV~(DJI Matrice 300 RTK) with Zenmuse H20T camera, which can achieve stable flight in extreme conditions, such as night, foggy, and windy scenes. The thermal camera captures 8-14$\mu m$ and we control the flight height from 5-20$m$ for a proper target size. We collect 500 sequences with 1,664,549 RGB-T image pairs. The images are in high quality, which are stored with 1920*1080 resolution\footnote{The thermal images are captured in 640*480 resolution and we rescale them to achieve registration in alignment process.} with jpg format and sampled in 30 fps. 
    We separate 250 sequences as training set and the other 250 sequences as test set. All sequences are also split into long-term and short-term sets, according to the absence of targets\footnote{We label the sequence as long-term, where the target is out-of-view for more than continuous 20 frames.}. In the training set, there are 207 short-term sequences and 43 long-term sequences. A total of 74 of 250 test sequences belong to long-term set to evaluate the performance for long-term tracking. Another 176 test sequences are used for short-term tracking evaluation. We also provide mask annotation for 100 sequences selected in short-term subset~(50 sequences for training and 50 sequences for test), which can be used for video object segmentation and scale estimation learning. Comparison between existing datasets
    can be found in Tab.~\ref{Statistics comparison} and Fig.~\ref{datasetDetail} (c).

    \begin{table*}[!t]
		\caption{Statistics comparison among existing multi-modality and UAV tracking datasets.}\label{Statistics comparison}
		\vspace{-3mm}
		\footnotesize
		\begin{center}
			\begin{tabular}{|p{0.5pt}<{\centering}lccccccccccc|}
			\hline
			& Benchmark & \makecell[c]{Num. \\ Seq.}  & \makecell[c]{Avg. \\ Frame}  & \makecell[c]{Min. \\ Frame} & \makecell[c]{Max. \\ Frame} & \makecell[c]{Total \\ Frame} & Resolution & \makecell[c]{Train \\ subset} & \makecell[c]{Long-term \\ subset}& \makecell[c]{Num. \\ Seg.} & Multi-modal & Year\\
			\hline
			\multirow{5}{*}{\rotatebox{90}{RGB-T}} & GTOT~\cite{Li_TIP16_GTOT} & 50 & 157 & 40 & 376 & 7.8K & 384 $\times$ 288 & \ding{53} &  \ding{53} & \ding{53} & \color{red} \ding{51} & 2016\\
			
			& RGBT210~\cite{Li_MM17_RGBT210} & 210 & 498 & 40 & 4140 & 104.7K & 630 $\times$ 460 & \ding{53} & \ding{53} & \ding{53} & \color{red} \ding{51} & 2017 \\
			
			& VOT2019-RGBT~\cite{Kristan_ICCVW19_VOT19} & 60 & 334 & 40 & 1335 & 40.2K & 630 $\times$ 460 & \ding{53} & \ding{53} & \ding{53} & \color{red} \ding{51} & 2019\\
			
			& RGBT234~\cite{Li_PR19_RGBT234} & 234 & 498 & 40 & 4140 & 116.7K & 630 $\times$ 460 & \ding{53} & \ding{53} & \ding{53} & \color{red} \ding{51} & 2019\\
			
			& LasHeR~\cite{Li_Arxiv21_LasHeR} & \bf 1224 & 600 & 57 & 12862 & 734.8K & 630 $\times$ 480 & \color{red} \ding{51} & \ding{53} & \ding{53} & \color{red} \ding{51} & 2021\\
			\hline
		    \multirow{3}{*}{\rotatebox{90}{UAV}} & UAV123~\cite{Mueller_ECCV16_UAV123} & 123 & 1246 & 109 & 5527 & 113.4K & 1280 $\times$ 720 &  \ding{53} & \color{red} \ding{51} & \ding{53} & \ding{53} & 2016\\
			& DTB~\cite{Li_AAAI17_VOT4UAV} & 70 & 225 & 68 & 699 & 15.7K & 1280 $\times$ 720 & \ding{53} & \ding{53} & \ding{53} & \ding{53} & 2017 \\
			& VisDrone-SOT~\cite{zhu2020visDrone} & 96 & 892 & 92 & 3135 & 85K & 1360 $\times$ 765 &\color{red}\ding{51} & $\times$ & \ding{53} & \ding{53} & 2018 \\
			\hline
			\rowcolor{gray!20} & \textbf{VTUAV} & 500 & \bf 3329 & \bf 196 & \bf 27213 & \bf 1.7M & {\bf 1920 $\times$ 1080} & \color{red}\ding{51} & \color{red}\ding{51} & \bf 24.4K & \color{red}\ding{51} & 2021\\
			\hline
			\end{tabular}
		\end{center}
	\end{table*}

    \item \emph{Generic object and scene category}.
    Related datasets are mainly recorded at the road, school for safely monitoring scenes, with limited scenes and object categories. We aim to construct a highly diverse dataset with adequate object types and scenes. As shown in Fig.~\ref{datasetDetail} (a), the tracked target can be divided into 5 super-classes~(pedestrian, vehicle, animal, train and ship) and 13 sub-classes, which can cover most of category for real-world applications. Sequences are captured at 15 scenes cross two cities, which include the road, street, bridge, park, sea, beach, court, and school, etc. To emphasize the effectiveness of both modalities, the data acquisition lasts for a whole year for various weather conditions and climates. Specifically, 325 sequences are captured at daytime and 175 sequences are at night with different conditions, such as windy, cloudy, and foggy weather.
    \item \emph{Hierarchical attributes}.
    Previous methods~\cite{Qi_AAAI19_ANT,Zhang_IJCV21_ADRNet,Li_ECCV20_CAT} aim to exploit the potential of attribute information and achieve satisfying performance in challenging cases. However, existing visual and RGB-T datasets~\cite{Wu_PAMI15_OTB100,Li_AAAI17_VOT4UAV,Li_PR19_RGBT234} label attributes at the sequence-level, which involves various challenges into single sequence in a coarse manner.
    In this paper, besides the sequence-wise attributes, we achieve a hierarchical attribute annotation by additionally labeling frame-level attributes for training sequences to fully investigate the attribute-based methods. Instead of separating the whole sequences into several clips~\cite{Fan_WACV2021_TracKlinic}, we maintain the sequence continuity, which allows the frames to be annotated with multiple labels or none.
    Challenges are summarized as 13 attributes, including target blur~(TB), camera movement~(CM), extreme illumination~(EI), deformation~(DEF), partial occlusion~(PO), full occlusion~(FO), scale variation~(SV), thermal crossover~(TC), fast moving~(FM), background clustering~(BC), out-of-view~(OV), low-resolution~(LR) and thermal-visible separation~(TVS).
    Fig.~\ref{datasetDetail} (b) lists the number of each attribute from sequence-level and frame-level. \emph{The description of attribute is summarized in supplementary material}. 
 
    \item \emph{Alignment}.
    Given that multi-sensor device cannot ensure the photocardic polymerization, thereby view differences occur. Previous RGB-T datasets apply frame-level alignment to calculate homography transformation and unify the view scope frame by frame, incurring immense labor costs and is impracticable in real-world applications. In our dataset, we operate modality alignment in the initial frame for each video and apply it to all frames. We note that most frames are well-aligned. \emph{The comparison of different alignment methods can be found in supplementary material.} 
\end{itemize} 

\subsection{High-quality Annotation}
In our dataset, we provide sufficient expert annotations in three formats, including bounding boxes, segmentation masks, and attribute annotations. Examples are shown in Fig.~\ref{fig:sample} and Fig.~\ref{datasetDetail} (d).
\begin{itemize}
    \item \emph{Bounding boxes}. In VTUAV, we carefully annotate bounding boxes for both modalities, individually. We provide sparse annotations in an interval of 10 frames. As described in~\cite{Muller_ECCV18_TrackingNet}, dense annotation can be achieved with the guidance of state-of-the-art trackers.
    In this manner, we provide 326,961 high-quality bounding box annotations in total.
    \item \emph{Segmentation masks}. We annotate the target mask at 1 fps for visible and thermal images. A total of 24,464 masks are generated using Labelme toolkit.
    
    \item \emph{Attribute annotations}. In our dataset, we provide frame-level attribute annotations to conduct the detailed attribute-based analysis. Most attributes\footnote{The attribute of FM, SV, LR and TVS are automatically annotated according to their descriptions.} are labeled by a full-time expert. Therefore, we totally label 301,678 frames with 430,960 attributes and provide 500 * 13 sequence-level annotations. 
\end{itemize}  

\subsection{Evaluation Metrics}
In our experiment, all the trackers are run in one-pass evaluation~(OPE) protocol and evaluated by maximum success rate (MSR) and maximum precision rate (MPR), which are widely used in RGB-T tracking~\cite{Li_TIP16_GTOT,Li_MM17_RGBT210,Li_PR19_RGBT234}. Overall performance for all sequences and attribute-based performance for attribute-specific sequences are considered. For mask evaluation, we measure the results with Jaccard index~($\cal J$) and F-score~($\cal F$)~\cite{Perazzi_CVPR16_DAVIS}.
\begin{itemize}
    \item \emph{Maximum success rate~(MSR)}. Success rate (SR) measures the ratio of tracked frames, determined by the Interaction-over-Union~(IoU) between tracking result and ground truth. With different overlap thresholds, a success plot~(SP) can be obtained, and SR is calculated as the area under curve of SP. Owing to the modality-level displacement, we adopt the maximum overlap in frame level as the final score.
    \item \emph{Maximum precision rate~(MPR)}. Similar to precision rate~(PR), MPR is to calculate the percentage of frames, where the center distance between prediction and ground truth is smaller than a threshold~$\tau$. $\tau$ is set to 20 in our experiment.
    \item \emph{Jaccard index~($\cal J$)}. Jaccard index is defined as the averaging pixel-level IoU between the predicted mask $\bf M$ and ground truth $\bf G$ for all $N$ frames. which can be formulated as ${\cal J} = \frac{1}{N} \sum_{i = 1}^{N} \frac{{\bf M}^i \cap {\bf G}^i}{{\bf M}^i \cup {\bf G}^i}$.
    
    \item \emph{F-score~($\cal F$)}. F-score calculates region-based precision $Pr$ and recall $Re$ based between the closed contours in $\bf M$ and $\bf G$, which can be expressed via ${\cal F} = \frac{2Pr*Re}{Pr+Re}$.

\end{itemize}
\begin{figure*}
    \centering
    \includegraphics{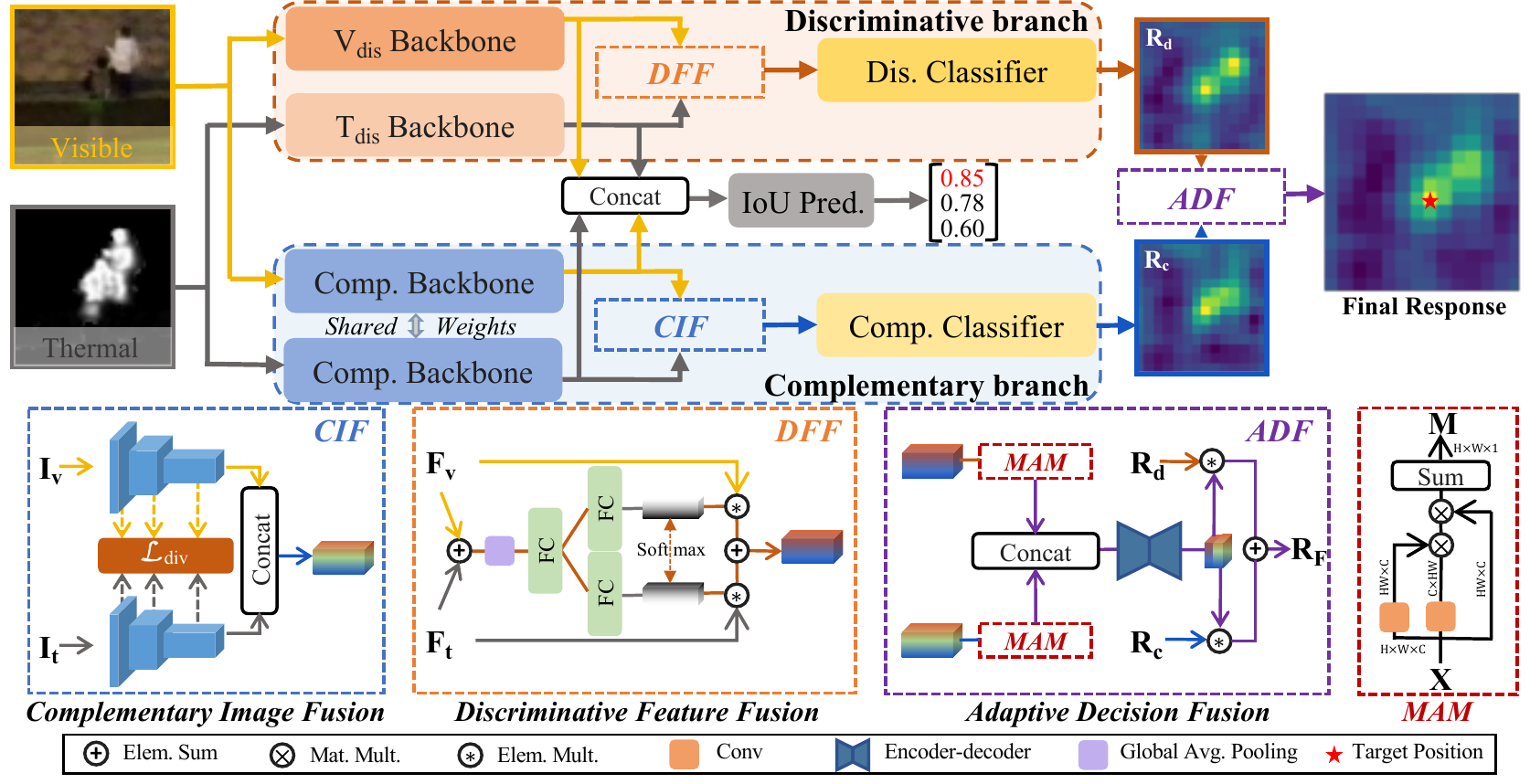}
    \caption{Overview of HMFT. Three fusion types are combined to learn a comprehensive representation and predict accurate results, which consists of Complementary Image Fusion~(CIF), Discriminative Feature Fusion~(DFF) and Adaptive Decision Fusion~(ADF). CIF aims to extract the modality-shared representation, while DFF fuses the individual features to learn the modality-independent map. Both two features are utilized to locate the target, and ADF is to make a final decision by combining the outputs of both branches.}
    \label{fig:framework}
\end{figure*}

\section{Hierarchical Multi-modal Fusion Tracker}
In this section, we will introduce a new baseline for RGB-T tracking
to fully exploit various fusion types in a unified framework, shown as Fig.~\ref{fig:framework}. It contains three main modules: Complementary Image Fusion~(CIF), Discriminative Feature Fusion~(DFF) and Adaptive Decision Fusion~(ADF). CIF aims to learn shared pattern between two modalities. DFF introduces a channel-wise combination of heterogeneous representations. Finally, ADF is to provide the final target candidate by considering the responses from discriminative and complementary classifiers.
\subsection{Complementary Image Fusion}
As shown in the third row in Fig.~\ref{fig:sample}, RGB-T images are captured in the same scenes and the complementary information~(such as semantic and contour, etc) exists in both modalities, which can be propagated to each other for a robust feature representation~\cite{Li_IF20_IVFuseNet}. 
To this end, we utilize a shared backbone, i.e., ResNet50~\cite{resnet}, to extract the common features. To leverage the consistency between the two modalities, we introduce a divergence loss ${\cal L}_{div}$ to constrain the multi-modal feature distribution by measuring their KL-divergence, which can be expressed as follows,

\begin{equation}
\begin{aligned}
    {\cal L}_{div} &= \sum_{i=1}^L \text{KL}({\bf P}^i_v||{\bf P}^i_t)\\
    &=\frac{1}{N}\sum_{i=1}^L\sum_{n=1}^N(p^i_{vn}\text{log}(p^i_{vn}-p^i_{tn}))
\end{aligned}
\end{equation}
where, ${\bf P}^i_v \in \mathbb{R}^{C*H*W}$ and ${\bf P}^i_t \in \mathbb{R}^{C*H*W}$ denote the complementary features output by $i$-th block for visible and thermal modalities, and $p^i_{vn}$ and $p^i_{tn}$ are the $n$-th items in ${\bf P}^i_v$ and ${\bf P}^i_t$, respectively. $L$ denotes the number of block in ResNet50. The learned representations are then concatenated to form the overall complementary feature ${\bf P}_a \in \mathbb{R}^{2C*H*W}$, where $C, H, W$ denote the channel number, height and width of the feature, respectively. Complementary feature is more robust when all modalities work well and achieves accurate scale estimation.
\subsection{Discriminative Feature Fusion}
Dual modalities can provide heterogeneous information, where visible images provide detailed context, and thermal images obtain more contour information according to the temperature difference, achieving robustness to illumination changes. To exploit the potential of both modalities, we first use an individual feature extractor to model each modality. Then we propose a Discriminative Feature Fusion~(DFF) module to fuse those representations. Considering the information from visible and thermal images, DFF provides a fused feature map by introducing a channel-wise modality weight. In DFF, feature maps from visible and thermal images ${\bf D}_v \in \mathbb{R}^{C*H*W}$ and ${\bf D}_t \in \mathbb{R}^{C*H*W}$ are summed, and we embed global vector ${\bf d}_g$ from both modalities by Global Average Pooling~(GAP) and Fully-Connected~(FC) layer, which can be expressed as,
\begin{equation}
    {\bf d}_{g} = {\cal F}_{g}(\text{GAP}({\bf D}_v + {\bf D}_t)),
\end{equation}
where $\cal F$($\cdot$) denotes the fully-connected layer. Then, two FC layers are adopted to produce channel-wise weights ${\bf w}_v, {\bf w}_t \in \mathbb{R}^{C*1*1}$ for each modality, which is followed by a softmax operation, as shown in Eq.~(\ref{weight_v}) and Eq.~(\ref{weight_t}).
\begin{equation}\label{weight_v}
    {\bf w}_v = \frac{e^{{\cal F}_v({\bf d}_g)}}{e^{{\cal F}_v({\bf d}_g)} + e^{{\cal F}_t({\bf d}_g)}}
\end{equation}
\begin{equation}\label{weight_t}
    {\bf w}_t = \frac{e^{{\cal F}_t({\bf d}_g)}}{e^{{\cal F}_v({\bf d}_g)} + e^{{\cal F}_t({\bf d}_g)}}
\end{equation}
Finally, the aggregated features can be obtained by weighted summation among channels via, 
\begin{equation}
{\bf D}^i_a = w^i_v * {\bf D}^i_v + w^i_t * {\bf D}^i_t
\end{equation}
where the superscript $i$ denotes the $i$-th channel of all the variables. With the proposed DFF, we construct a comprehensive feature, which fuses the latent representations of visible-thermal modalities.

\subsection{Adaptive Decision Fusion}
The aforementioned CIF and DFF make decisions individually. They model complementary and discriminative cues, respectively, which output as two response maps. It is crucial to determine which cue is reliable for target location. Thus, we introduce Adaptive Decision Fusion~(ADF) to fuse these two response maps according to their modality confidences. First, the Modality Aggregation Module~(MAM) is designed to obtain the confidence for each modality. MAM is a self-attention network, which produces the modality confidence ${\bf M}_d$ and ${\bf M}_c$. It mines the modality information in a non-local manner, which takes the whole coordinates into consideration. The MAM process can be formulated as, 
\begin{equation}
{\bf M} = {\cal S}({\bf A} \times {\bf X}),
\end{equation}
\begin{equation}
{\bf A} = {\cal R}(\phi_{1*1}({\bf X})) \times {\cal R}(\varphi_{1*1}({\bf X})),
\end{equation}
where $\bf X$ is the input feature, i.e., ${\bf P}_a$ or ${\bf D}_a$. $\phi_{1*1}(\cdot)$ and $\varphi_{1*1}(\cdot)$ are the learnable $1*1$ Convolution layers. ${\cal R}(\cdot)$ and ${\cal S}(\cdot)$ are the reshape operation and channel-wise summation, respectively. $\times$ denotes matrix multiplication. When the modality confidences for discriminative and complementary branches ${\bf M}_d$ and ${\bf M}_c$ are calculated, they are concatenated and sent to a two-layer encoder-decoder network to generate the weight maps ${\bf E}_d \in {\mathbb R}^{H*W}$ and ${\bf E}_c \in {\mathbb R}^{H*W}$  for the response. Finally, the final response is obtained by ${\bf R}_F = {\bf R}_d \odot {\bf E}_d + {\bf R}_c \odot {\bf E}_c$, where $\odot$ denotes element-wise production.

\subsection{Implementation Details}
We take DiMP~\cite{Bhat_ICCV19_DIMP} as our base tracker, with the truncated ResNet50 as the backbone network. We conduct a multi-step training process to fit different purposes in various modules. First, we train both backbones for discriminative and complementary branches. The losses for both branches are expressed as follows,
\begin{equation}
\label{loss}
\begin{aligned}
&{\cal L}_d = {\cal L}_{bb} + \beta {\cal L}_{cls}\\
&{\cal L}_c = {\cal L}_{bb} + \beta {\cal L}_{cls} + \gamma {\cal L}_{div}
\end{aligned}
\end{equation}
where, ${\cal L}_{bb}$ and ${\cal L}_{cls}$ are the bounding box estimation and target classification losses, respectively, which are detailed in DiMP~\cite{Bhat_ICCV19_DIMP}. $\beta$ and $\gamma$ are the weights for classification and divergence loss, which are set to 100.
After that, the learned backbones are fixed, and we learn DFF module and the classifiers with Eq.~(\ref{loss}). Finally, with all the backbones fixed, we start to learn ADF and IoU prediction modules~\cite{Jiang_ECCV18_IoUNet} and fine-tune both classifiers to fit the learned representations. The learning rates for DFF and ADF are $2e^{-5}$ and $2e^{-4}$. We use the same setting in DiMP~\cite{Zhang_ICCVW19_DIMP-RGBT} to train the backbone. We fine-tune the network with multiplying a decreasing factor to the original learning rate, which is set to 0.1.
HMFT is implemented on Pytorch platform and run on a single Nvidia RTX Titan GPU with 24G memory.

\begin{table*}[t]
	\caption{Comparison results for short-term tracking on the proposed dataset and existing RGB-T tracking benchmarks, including GTOT, RGBT210 and RGBT234. The top-three trackers are marked in {\color{red}red}, {\color{blue}blue} and {\color{green}green} fonts.}

	\label{tab:SOTA_ST}
	\footnotesize
	\begin{center}
		\begin{tabular}{l|cc|cc|cc|cc|c}
			\hline
			\multirow{2}{*}{Tracker} &
			\multicolumn{2}{c|}{VTUAV} &
			\multicolumn{2}{c|}{GTOT} & \multicolumn{2}{c|}{RGBT210} & \multicolumn{2}{c|}{RGBT234} & \multirow{2}{*}{FPS}\\
			& MSR & MPR & MSR & MPR & MSR & MPR & MSR & MPR & \\
			\hline
			DAFNet~\cite{Gao_ICCVW19_DAFNet} & 45.8 & 62.0 & \color{green} 71.2 & \color{green} 89.1 & 48.5 & 72.6 & \color{green} 54.4 & \color{blue} 79.6 & 21.0\\
			ADRNet~\cite{Zhang_IJCV21_ADRNet} & 46.6 & 62.2 & \color{blue} 73.9 & \color{blue} 90.4 & \color{blue} 53.4 & \color{blue} 77.8 & \color{red} 57.1 & \color{red} 80.9 & 25.0\\
			FSRPN ~\cite{Kristan_ICCVW19_VOT19} & \color{green} 54.4 & \color{green} 65.3 & 69.5 & 89.0 & 49.6 & 68.9 & 52.5 & 71.9 & \color{red} 30.3\\
			mfDiMP~\cite{Zhang_ICCVW19_DIMP-RGBT} & \color{blue} 55.4 & \color{blue} 67.3 & 49.0 & 59.4 & \color{green} 52.2 & \color{green} 74.9 & 42.8 & 64.6 & \color{green} 28.0\\
			\rowcolor{gray!20} HMFT~(Ours) & \color{red} 62.7 & \color{red} 75.8 & \color{red} 74.9 & \color{red} 91.2 & \color{red} 53.5 & \color{red} 78.6 & \color{blue} 56.8 & \color{green} 78.8 & \color{blue} 30.2\\
			\hline
		\end{tabular}
	\vspace{-5mm}
	\end{center}

\end{table*}

\section{Experimental Analysis for RGB-T Tracking}
\subsection{Short-term Evaluation}
\noindent {\bf Overall performance}. We select four RGB-T trackers~(DAFNet~\cite{Gao_ICCVW19_DAFNet}, ADRNet~\cite{Zhang_IJCV21_ADRNet}, FSRPN~\cite{Kristan_ICCVW19_VOT19} and mfDiMP~\cite{Zhang_ICCVW19_DIMP-RGBT}). As shown in Tab.~\ref{tab:SOTA_ST}, HMFT with real-time speed achieves the top performance with 62.7\% MSR and 75.8\% MPR. The runner-up tracker is mfDiMP, which equips a box regression module to apply scale estimation. Siamese-based tracker~(FSRPN) and multi-domain networks~(DAFNet and ADRNet) obtain inferior results.\\
\noindent {\bf Comparison on existing datasets}. We also conduct analysis on three popular RGB-T tracking benchmarks, including GTOT, RGBT210 and RGBT234. To adapt on different benchmarks, we fine-tune HMFT on GTOT and test on RGBT210 and RGBT234. The results are shown in Tab.~\ref{tab:SOTA_ST}. HMFT obtains satisfying performance in all public benchmarks with real-time  speed. Specifically, HMFT achieves the state-of-the-art performance in GTOT and RGBT210, with 74.9\% and 53.5\% MSR and 91.2\% and 78.6\% MPR. In RGBT234, our tracker obtains the top-three results against all the competitors with 56.8\% and 78.8\% in MSR and MPR. All the results show the effectiveness of HMFT, which has great potential for a strong baseline tracker.

\subsection{Long-term Evaluation}
\noindent{\bf Overall performance}. Given no available long-term RGB-T trackers, following the idea of \cite{Dai_CVPR20_LTMU}, we implement the HMFT\_LT with a global tracker, in which we utilize GlobalTrack~\cite{Huang_AAAI20_GlobalTrack} as a global detector and RTMDNet~\cite{Jung_ECCV18_RT-MDNet} as the tracker switcher. When RTMDNet recognizes the absence of target, GlobalTrack is selected to find the target in the whole images. We test all the competitors and the results are shown in Tab.~\ref{SOTA_LT}. Our long-term variant~(HMFT\_LT) sets a new baseline for RGB-T long-term tracking, which outperforms short-term verison~(HMFT) with 29.8\% and 29.4\% relative promotion in MSR and MPR, respectively.
\begin{table}[h]
	\caption{Quantitative comparison against state-of-the-art RGB-T trackers on the long-term subset.}

	\label{SOTA_LT}
	\small
	\begin{center}
		\vspace{-3mm}
		\begin{tabular}{lccc}
		    \hline
			Tracker & MSR & MPR & FPS\\
			\hline
			ADRNet~\cite{Zhang_IJCV21_ADRNet} & 17.5 & 23.5 & 10.3\\
			DAFNet~\cite{Gao_ICCVW19_DAFNet} & 18.8 & 25.3 & 7.1\\
			mfDiMP~\cite{Zhang_ICCVW19_DIMP-RGBT} & 27.2 & 31.5 & \color{blue} 25.8\\
			FSRPN~\cite{Kristan_ICCVW19_VOT19} & \color{green} 31.4 & \color{green} 36.6 & \color{red} 36.8\\
			\rowcolor{gray!20}\bf HMFT~(Ours) & \color{blue} 35.5 & \color{blue} 41.4 & \color{green} 25.1\\
			\rowcolor{gray!20}\bf HMFT\_LT~(Ours) & \color{red} 46.1 & \color{red} 53.6 & 8.1\\
			\hline
		\end{tabular}
	\end{center}
	\vspace{-3mm}
\end{table}

\subsection{Ablation Study}
\begin{figure}[t]
    \centering
    \includegraphics[width=0.45\textwidth]{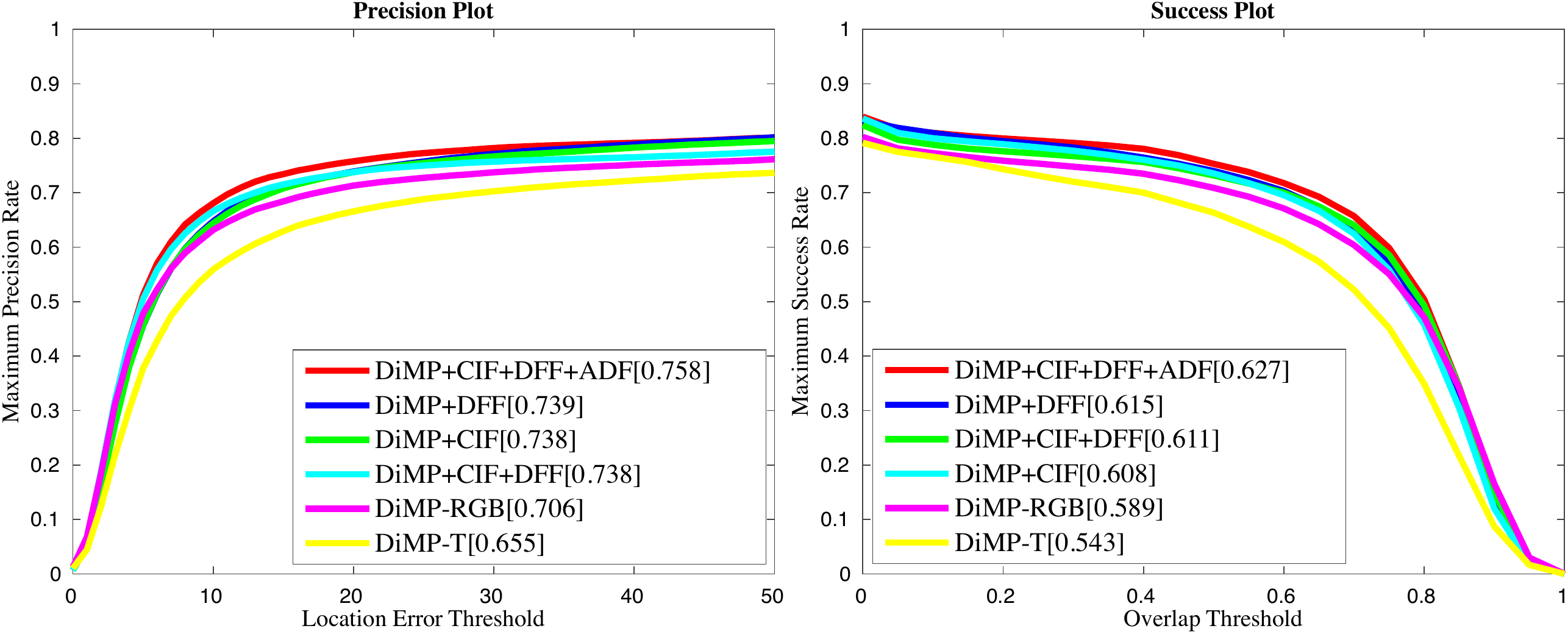}
    \caption{Ablation analysis of HMFT.}
    \label{fig:ablation}
    \vspace{-5mm}
\end{figure}

The ablation analysis of HMFT is shown in Fig.~\ref{fig:ablation}. Given that visible images contain more detailed information, which is more capable of recognizing the object, DiMP with the visible image is much more superior to that with the thermal image. Furthermore, both image fusion and feature fusion achieve obvious promotion against trackers with single modality~(DiMP-RGB and DiMP-T), where DiMP+CIF and DiMP+DFF gain 3.2\% and 4.4\% promotion in MSR, respectively. DiMP+CIF+DFF simply averages the responses from complementary and discriminative branches, leading to a slight performance decrease. Our final model~(DiMP+CIF+DFF+ADF) achieves 2.6\% and 2.7\% improvement in MSR and MPR, indicating the adaptation of our decision fusion module.

\subsection{Qualitative Analysis}
\begin{figure}[t]
    \centering
    \includegraphics[width=0.45\textwidth]{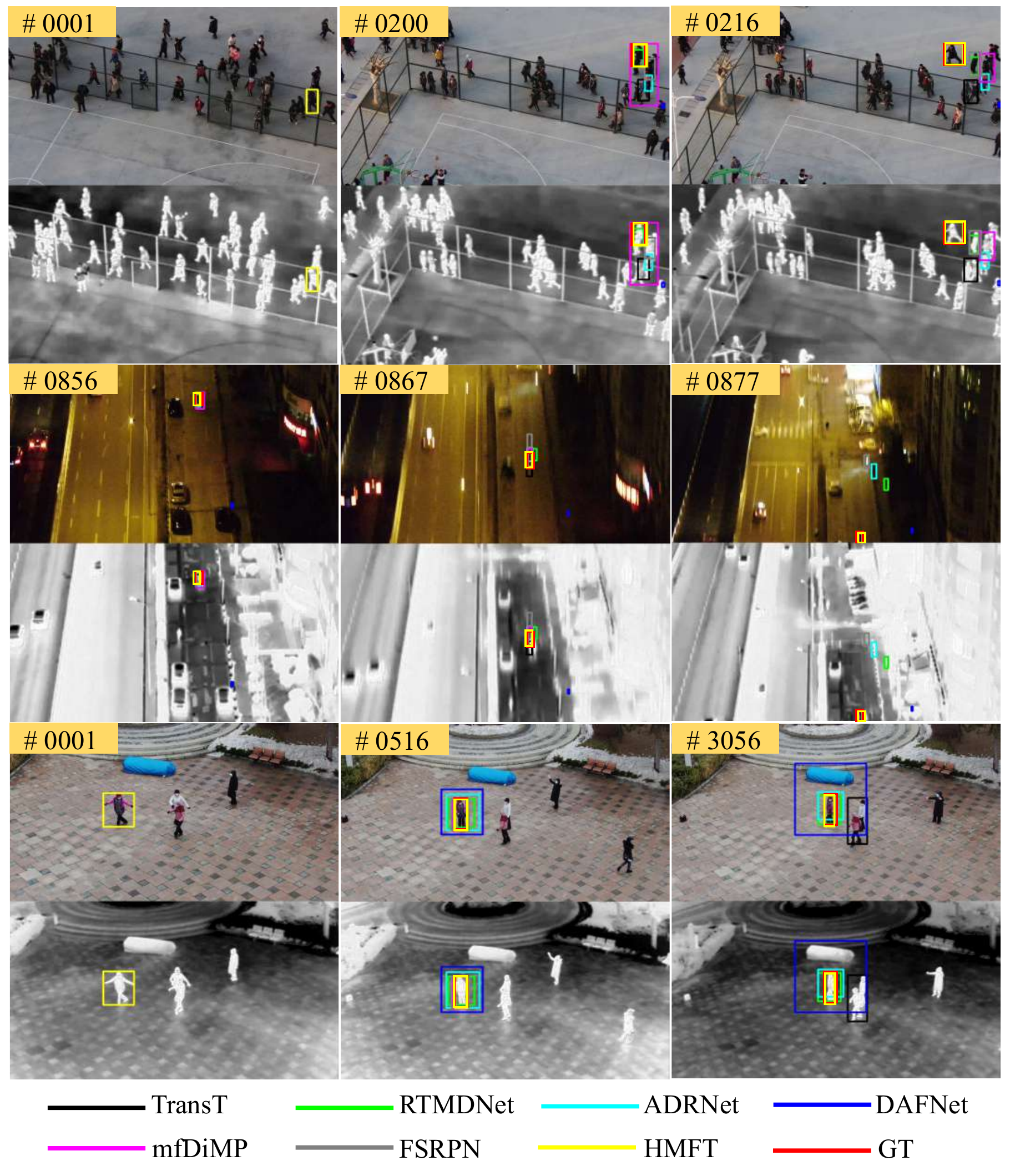}
    \caption{Qualitative comparison of HMFT. Our method shows strong performance on PO, CM and SV.}
    \label{fig:QA}
    \vspace{-4mm}
\end{figure}

Fig.~\ref{fig:QA} provides visualization results between HMFT and the competitors. HMFT shows accurate tracking results on various challenges, such as occlusion, camera movement and scale variation, while other trackers miss the target or cannot estimate the scale properly.
\section{Experimental Results on VTUAV-V Subset }
We state that VTUAV has great potential for conventional visual tracking task. To unveil the power of RGB tracking, we construct a subset, namely VTUAV-V, which only contains the visible modality for RGB tracking. We evaluate numerous popular short-term and long-term RGB trackers with various frameworks on VTUAV-V dataset, including, LTMU~\cite{Dai_CVPR20_LTMU}, STARK~\cite{Yan_ICCV21_STARK}, TransT~\cite{Chen_CVPR21_TransT}, DiMP\cite{Bhat_ICCV19_DIMP}, SiamRPN++~\cite{Li_CVPR19_SiamRPN++}, ATOM~\cite{Martin_CVPR19_ATOM}, LightTrack\cite{Yan_CVPR21_LightTrack}, Ocean~\cite{Zhang_ECCV20_Ocean}, SiamRPN~\cite{Li_CVPR18_SiamRPN}, SiamTPN~\cite{Xing_WACV22_SiamTPN}, SiamAPN++\cite{Cao_IROS21_SiamAPN++}, ECO~\cite{Danelljan_CVPR17_ECO}, D3S~\cite{lukezic_CVPR2020_d3s}, RTMDNet~\cite{Jung_ECCV18_RT-MDNet}, HiFT~\cite{Cao_ICCV21_HiFT}, SPLT~\cite{Yan_ICCV19_SPLT}, SiamFC~\cite{Bertinetto_ECCVW16_SiamFC} and GlobalTrack~\cite{Huang_AAAI20_GlobalTrack}. Since only visible modality is used for tracking. We measure their performance using SR and PR instead of calculating the maximum score between those two modalities. Each tracker is tested without any modification or retraining.
\subsection{Short-term Evaluation}
\begin{figure}[t]
    \centering
    \includegraphics[width=0.45\textwidth]{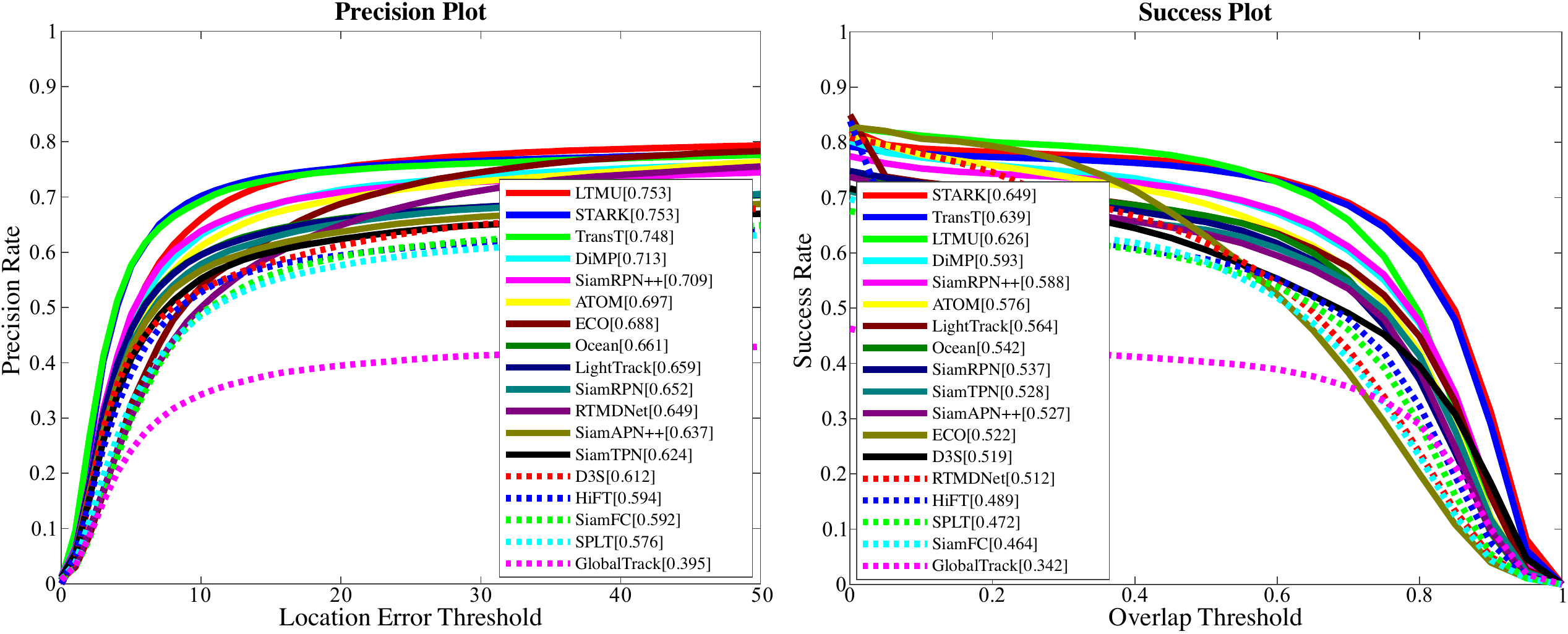}
    \caption{Evaluation results on short-term subset of VTUAV-V.}
    \label{fig:RGB_SOTA_ST}
\end{figure}

As shown in Fig.~\ref{fig:RGB_SOTA_ST}, transformer-based trackers~(STARK, TransT) obtain the top performances, which shows the dominative strength in tracking. STARK achieves the top performance with 64.9\% and 75.3\% in SR and PR, respectively. With the global tracker, LTMU can redetect the target when suffering large camera motion and view-point change, leading to a satisfying performance with respect to PR. Trackers with online updating~(LTMU, DiMP, ATOM, ECO) obtain the following-up performances and Siamese-based trackers~(SiamRPN++, LightTrack, Ocean, SiamRPN, SiamTPN, SiamAPN++, HiFT) achieve inferior results, which indicates the importance of model updating.
\begin{figure}[t]
    \centering
    \includegraphics[width=0.45\textwidth]{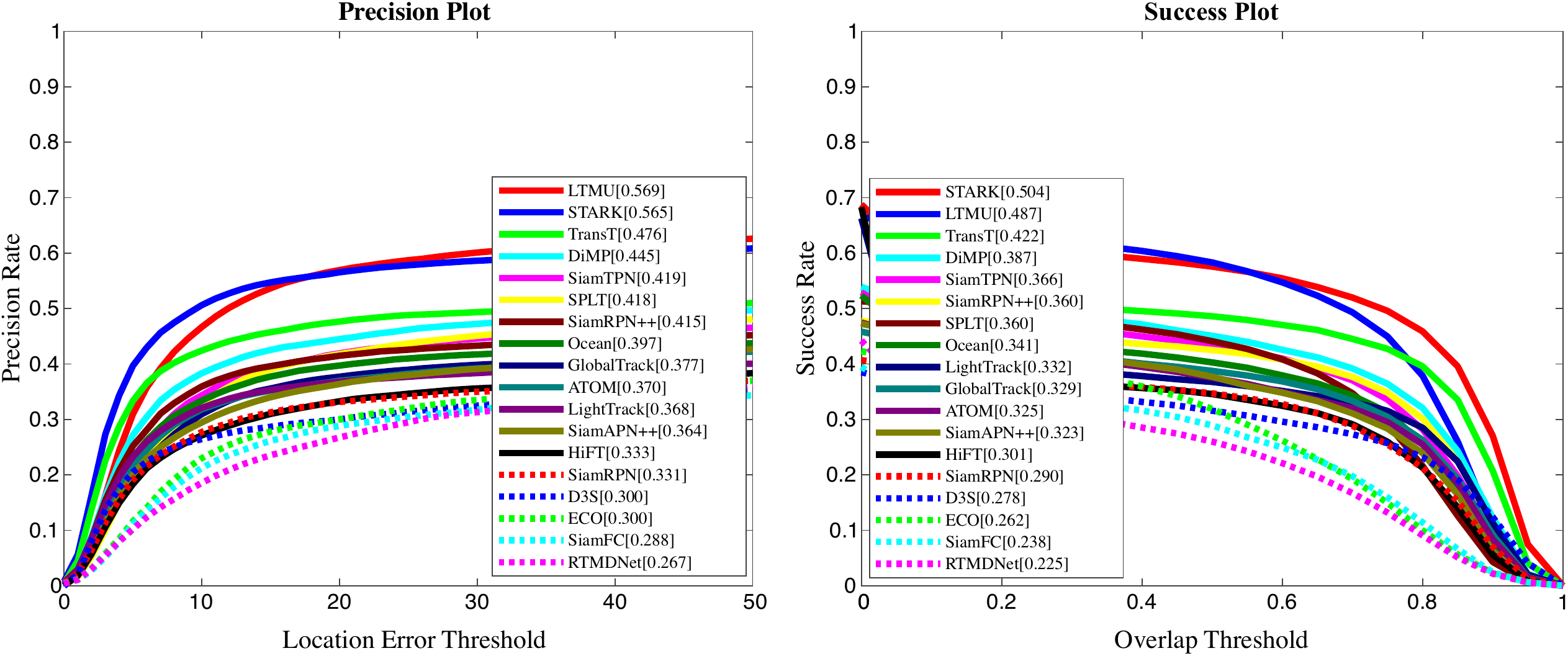}
    \caption{Evaluation results on long-term subset of VTUAV-V.}
    \label{fig:RGB_SOTA_LT}
    \vspace{-3mm}
\end{figure}

\subsection{Long-term Evaluation}
As shown in Fig.~\ref{fig:RGB_SOTA_LT}, compared with the performances on short-term subset, all the trackers show an decreasing performance. STARK and LTMU obtain the top performance on SR and PR, respectively. With the global detection mechanism, all long-term trackers~(LTMU, GlobalTrack and SPLT) promote their rankings signicantly due to their redetection modules.

\section{Conclusion}
\vspace{2mm}
In this paper, we release a large-scale benchmark for RGB-T tracking. Three main breakthroughs have been achieved. First, we address the issue of few available training data by providing diverse and high-resolution paired RGB-T images captured in various conditions. Second, to the best of our knowledge, this is the first unified RGB-T dataset that considers short-term tracking, long-term tracking and pixel-level prediction to evaluate the trackers comprehensively. Third, we annotate 13 challenges in sequence and frame levels, which can meet the requirement of scene-specific trackers and fully exploit the effectiveness of attributes. Moreover, a new baseline, called HFMT, is designed by combining both image fusion, feature fusion and decision fusion. The remarkable performance on three benchmarks show the complement of those fusion methods, and the importance of adequate training data.
\vspace{4mm}

\clearpage
{\small
\bibliographystyle{ieee_fullname}
\bibliography{egbib}

\begin{thebibliography}{10}\itemsep=-1pt

\bibitem{Bertinetto_ECCVW16_SiamFC}
Luca Bertinetto, Jack Valmadre, Joao~F. Henriques, Andrea Vedaldi, and Philip
  H.~S. Torr.
\newblock Fully-convolutional siamese networks for object tracking.
\newblock In {\em European Conference on Computer Vision Workshop}, pages
  850--865, 2016.

\bibitem{Bhat_ICCV19_DIMP}
Goutam Bhat, Martin Danelljan, Luc~Van Gool, and Radu Timofte.
\newblock Learning discriminative model prediction for tracking.
\newblock In {\em IEEE International Conference on Computer Vision}, pages
  6182--6191, 2019.

\bibitem{Cao_ICCV21_HiFT}
Ziang Cao, Changhong Fu, Junjie Ye, Bowen Li, and Yiming Li.
\newblock Hift: Hierarchical feature transformer for aerial tracking.
\newblock In {\em IEEE International Conference on Computer Vision}, pages
  15457--15466, 2021.

\bibitem{Cao_IROS21_SiamAPN++}
Ziang Cao, Changhong Fu, Junjie Ye, Bowen Li, and Yiming Li.
\newblock Siamapn++: Siamese attentional aggregation network for real-time
  {UAV} tracking.
\newblock In {\em International Conference on Intelligent Robots and Systems},
  pages 3086--3092, 2021.

\bibitem{Chen_CVPR21_TransT}
Xin Chen, Bin Yan, Jiawen Zhu, Dong Wang, Xiaoyun Yang, and Huchuan Lu.
\newblock Transformer tracking.
\newblock In {\em IEEE Conference on Computer Vision and Pattern Recognition},
  pages 8126--8136, 2021.

\bibitem{Dai_CVPR20_LTMU}
Kenan Dai, Yunhua Zhang, Dong Wang, Jianhua Li, Huchuan Lu, and Xiaoyun Yang.
\newblock High-performance long-term tracking with meta-updater.
\newblock In {\em IEEE Conference on Computer Vision and Pattern Recognition},
  pages 6298--6307, 2020.

\bibitem{Danelljan_CVPR17_ECO}
Martin Danelljan, Goutam Bhat, Fahad~Shahbaz Khan, and Michael Felsberg.
\newblock {ECO}: Efficient convolution operators for tracking.
\newblock In {\em IEEE Conference on Computer Vision and Pattern Recognition},
  pages 6638--6646, 2017.

\bibitem{Martin_CVPR19_ATOM}
Martin Danelljan, Goutam Bhat, Fahad~Shahbaz Khan, and Michael Felsberg.
\newblock {ATOM}: Accurate tracking by overlap maximization.
\newblock In {\em IEEE Conference on Computer Vision and Pattern Recognition},
  pages 4660--4669, 2019.

\bibitem{Fan_WACV2021_TracKlinic}
Heng Fan, Fan Yang, Peng Chu, Yuewei Lin, Lin Yuan, and Haibin Ling.
\newblock {TracKlinic}: Diagnosis of challenge factors in visual tracking.
\newblock In {\em IEEE Winter Conference on Applications of Computer Vision},
  pages 970 -- 979, 2021.

\bibitem{Gao_ICCVW19_DAFNet}
Yuan Gao, Chenglong Li, Yabin Zhu, Jin Tang, Tao He, and Futian Wang.
\newblock Deep adaptive fusion network for high performance {RGBT} tracking.
\newblock In {\em IEEE International Conference on Computer Vision Workshop},
  pages 1--9, 2019.

\bibitem{resnet}
Kaiming He, Xiangyu Zhang, Shaoqing Ren, and Jian Sun.
\newblock Deep residual learning for image recognition.
\newblock In {\em IEEE Conference on Computer Vision and Pattern Recognition},
  pages 770--778, 2016.

\bibitem{Huang_AAAI20_GlobalTrack}
Lianghua Huang, Xin Zhao, and Kaiqi Huang.
\newblock {GlobalTrack}: A simple and strong baseline for long-term tracking.
\newblock In {\em AAAI Conference on Artificial Intelligence}, pages
  11037--11044, 2020.

\bibitem{Jiang_ECCV18_IoUNet}
Borui Jiang, Ruixuan Luo, Jiayuan Mao, Tete Xiao, and Yuning Jiang.
\newblock Acquisition of localization confidence for accurate object detection.
\newblock In {\em European Conference on Computer Vision}, pages 784--799,
  2018.

\bibitem{Jung_ECCV18_RT-MDNet}
Ilchae Jung, Jeany Son, Mooyeol Baek, and Bohyung Han.
\newblock Real-time {MDN}et.
\newblock In {\em European Conference on Computer Vision}, pages 83--98, 2018.

\bibitem{Kristan_ICCVW19_VOT19}
Matej Kristan, Jiri Matas, Ales Leonardis, Michael Felsberg, and et al.
\newblock The seventh visual object tracking {VOT2019} challenge results.
\newblock In {\em IEEE International Conference on Computer Vision Workshop},
  pages 1--36, 2019.

\bibitem{Kristan_ECCVW20_VOT20}
Matej Kristan, Jiri Matas, Ales Leonardis, Michael Felsberg, and et al.
\newblock The eighth visual object tracking {VOT2020} challenge results.
\newblock In {\em European Conference on Computer Vision Workshop}, pages
  547--601, 2020.

\bibitem{Li_CVPR19_SiamRPN++}
Bo Li, Wei Wu, Qiang Wang, Fangyi Zhang, Junliang Xing, and Junjie Yan.
\newblock Siamrpn++: Evolution of siamese visual tracking with very deep
  networks.
\newblock In {\em IEEE Conference on Computer Vision and Pattern Recognition},
  pages 4282--4291, 2019.

\bibitem{Li_CVPR18_SiamRPN}
Bo Li, Junjie Yan, Wei Wu, Zheng Zhu, and Xiaolin Hu.
\newblock High performance visual tracking with siamese region proposal
  network.
\newblock In {\em IEEE Conference on Computer Vision and Pattern Recognition},
  pages 8971--8980, 2018.

\bibitem{Li_TIP16_GTOT}
Chenglong Li, Hui Cheng, Shiyi Hu, Xiaobai Liu, Jin Tang, and Liang Lin.
\newblock Learning collaborative sparse representation for grayscale-thermal
  tracking.
\newblock {\em IEEE Transactions on Image Processing}, 25(12):5743--5756, 2016.

\bibitem{Li_PR19_RGBT234}
Chenglong Li, Xinyan Liang, Yijuan Lu, Nan Zhao, and Jin Tang.
\newblock {RGB-T} object tracking: Benchmark and baseline.
\newblock {\em Pattern Recognition}, 96(12):106977, 2019.

\bibitem{Li_ECCV20_CAT}
Chenglong Li, Lei Liu, Andong Lu, Qing Ji, and Jin Tang.
\newblock Challenge-aware {RGBT} tracking.
\newblock In {\em European Conference on Computer Vision}, pages 222--237,
  2020.

\bibitem{Li_ICCVW19_MANet}
Chenglong Li, Andong Lu, Aihua Zheng, Zhengzheng Tu, and Jin Tang.
\newblock Multi-adapter {RGBT} tracking.
\newblock In {\em IEEE International Conference on Computer Vision Workshop},
  pages 2262--2270, 2019.

\bibitem{Li_NEU18_FTSNet}
Chenglong Li, Xiaohao Wu, Nan Zhao, Xiaochun Cao, and Jin Tang.
\newblock Fusing two-stream convolutional neural networks for {RGB-T} object
  tracking.
\newblock {\em Neurocomputing}, 281:78--85, 2018.

\bibitem{Li_Arxiv21_LasHeR}
Chenglong Li, Wanlin Xue, Yaqing Jia, Zhichen Qu, Bin Luo, and Jin Tang.
\newblock {LasHeR}: A large-scale high-diversity benchmark for {RGBT} tracking.
\newblock {\em arXiv preprint arXiv:2104.13202}, 2021.

\bibitem{Li_MM17_RGBT210}
Chenglong Li, Nan Zhao, Yijuan Lu, Chengli Zhu, and Jin Tang.
\newblock Weighted sparse representation regularized graph learning for {RGB-T}
  object tracking.
\newblock In {\em ACM International Conference on Multimedia}, pages
  1856--1864, 2017.

\bibitem{Li_AAAI17_VOT4UAV}
Siyi Li and Dit-Yan Yeung.
\newblock Visual object tracking for unmanned aerial vehicles: A benchmark and
  new motion models.
\newblock In {\em AAAI Conference on Artificial Intelligence}, pages 4140 --
  4146, 2017.

\bibitem{Li_IF20_IVFuseNet}
Yuqi Li, Haitao Zhao, Zhengwei Hu, Qianqian Wang, and Yuru Chen.
\newblock {IVFuseNet}: Fusion of infrared and visible light images for depth
  prediction.
\newblock {\em Information Fusion}, 58:1 -- 12, 2020.

\bibitem{Tracking-Book}
Huchuan Lu and Dong Wang.
\newblock {\em Online Visual Tracking}.
\newblock Springer, 2019.

\bibitem{lukezic_CVPR2020_d3s}
Alan Lukezic, Jiri Matas, and Matej Kristan.
\newblock {D3S-A} discriminative single shot segmentation tracker.
\newblock In {\em IEEE Conference on Computer Vision and Pattern Recognition},
  pages 7133--7142, 2020.

\bibitem{Luo_IPT19_AWS}
Chengwei Luo, Bin Sun, Ke Yang, Taoran Lu, and Wei-Chang Yeh.
\newblock Thermal infrared and visible sequences fusion tracking based on a
  hybrid tracking framework with adaptive weighting scheme.
\newblock {\em Infrared Physics and Technology}, 99:265--276, 2019.

\bibitem{Mueller_ECCV16_UAV123}
Matthias Mueller, Neil Smith, and Bernard Ghanem.
\newblock A benchmark and simulator for {UAV} tracking.
\newblock In {\em European Conference on Computer Vision}, pages 445--461,
  2016.

\bibitem{Muller_ECCV18_TrackingNet}
Matthias Muller, Adel Bibi, Silvio Giancola, Salman Alsubaihi, and Bernard
  Ghanem.
\newblock Tracking{N}et: A large-scale dataset and benchmark for object
  tracking in the wild.
\newblock In {\em European Conference on Computer Vision}, pages 300 -- 317,
  2018.

\bibitem{Peng_arxiv21_SiamIVFN}
Jingchao Peng, Haitao Zhao, Zhengwei Hu, Yi Zhuang, and Bofan Wang.
\newblock Siamese infrared and visible light fusion network for {RGB-T}
  tracking.
\newblock {\em arXiv preprint arXiv:2103.07302}, 2021.

\bibitem{Perazzi_CVPR16_DAVIS}
Federico Perazzi, Jordi Pont-Tuset, Brian McWilliams, Luc~Van Gool, Markus
  Gross, and Alexander Sorkine-Hornung.
\newblock A benchmark dataset and evaluation methodology for video object
  segmentation.
\newblock In {\em IEEE Conference on Computer Vision and Pattern Recognition},
  pages 724--732, 2016.

\bibitem{Qi_AAAI19_ANT}
Yuankai Qi, Shengping Zhang, Weigang Zhang, Li Su, Qingming Huang, and
  Ming-Hsuan Yang.
\newblock Learning attribute-specific representations for visual tracking.
\newblock In {\em AAAI Conference on Artificial Intelligence}, pages
  8835--8842, 2019.

\bibitem{Torabi_CVIU12_LITIV}
Atousa Torabi, Guillaume Massé, and Guillaume-Alexandre Bilodeau.
\newblock An iterative integrated framework for thermal-visible image
  registration, sensor fusion, and people tracking for video surveillance
  applications.
\newblock {\em Computer Vision and Image Understanding}, 116(2):210--221, 2012.

\bibitem{Wang_CVPR20_CMPP}
Chaoqun Wang, Chunyan Xu, Zhen Cui, Ling Zhou, Tong Zhang, Xiaoya Zhang, and
  Jian Yang.
\newblock Cross-modal pattern-propagation for {RGB-T} tracking.
\newblock In {\em IEEE Conference on Computer Vision and Pattern Recognition},
  pages 7064--7073, 2020.

\bibitem{Wang_arxiv2021_MGFNet}
Xiao Wang, Xiujun Shu, Shiliang Zhang, Bo Jiang, Yaowei Wang, Yonghong Tian,
  and Feng Wu.
\newblock {MFGNet}: Dynamic modality-aware filter generation for {RGB-T}
  tracking.
\newblock {\em arXiv preprint arXiv:2107.10433}, 2021.

\bibitem{Davis_CVIU07_OTCBVS}
James W.Davis and Vinay Sharma.
\newblock Background-subtraction using contour-based fusion of thermal and
  visible imagery.
\newblock {\em Computer Vision and Image Understanding}, 106(2-3):162--182,
  2007.

\bibitem{Wu_PAMI15_OTB100}
Yi Wu, Jongwoo Lim, and Ming-Hsuan Yang.
\newblock Object tracking benchmark.
\newblock {\em IEEE Transactions on Pattern Analysis and Machine Intelligence},
  37(9):1834--1848, 2015.

\bibitem{Xing_WACV22_SiamTPN}
Daitao Xing, Nikolaos Evangeliou, Athanasios Tsoukalas, and Anthony Tzes.
\newblock Siamese transformer pyramid networks for real-time {UAV} tracking.
\newblock In {\em IEEE Winter Conference on Applications of Computer Vision},
  pages 1898--1907, 2022.

\bibitem{Yan_ICCV21_STARK}
Bin Yan, Houwen Peng, Jianlong Fu, Dong Wang, and Huchuan Lu.
\newblock Learning spatio-temporal transformer for visual tracking.
\newblock In {\em IEEE International Conference on Computer Vision}, pages
  10428--10437, 2021.

\bibitem{Yan_CVPR21_LightTrack}
Bin Yan, Houwen Peng, Kan Wu, Dong Wang, Jianlong Fu, and Huchuan Lu.
\newblock Lighttrack: Finding lightweight neural networks for object tracking
  via one-shot architecture search.
\newblock In {\em IEEE Conference on Computer Vision and Pattern Recognition},
  pages 15180--15189, 2021.

\bibitem{Yan_ICCV19_SPLT}
Bin Yan, Haojie Zhao, Dong Wang, Huchuan Lu, and Xiaoyun Yang.
\newblock `skimming-perusal' tracking: A framework for real-time and robust
  long-term tracking.
\newblock In {\em IEEE International Conference on Computer Vision}, pages
  2385--2393, 2019.

\bibitem{Zhang_Sensor20_MaCNet}
Hui Zhang, Lei Zhang, Li Zhuo, and Jing Zhang.
\newblock Object tracking in {RGB-T} videos using modal-aware attention network
  and competitive learning.
\newblock {\em Sensors}, 20(2):1--19, 2020.

\bibitem{Zhang_ICCVW19_DIMP-RGBT}
Lichao Zhang, Martin Danelljan, Abel Gonzalez-Garcia1, Joost van~de Weijer, and
  Fahad~Shahbaz Khan.
\newblock Multi-modal fusion for end-to-end {RGB-T} tracking.
\newblock In {\em IEEE International Conference on Computer Vision Workshop},
  pages 1--10, 2019.

\bibitem{Zhang_Arxiv20_MM_tracking_survey}
Pengyu Zhang, Dong Wang, and Huchuan Lu.
\newblock Multi-modal visual tracking: Review and experimental comparison.
\newblock {\em arXiv preprint arXiv:2012.04176}, 2020.

\bibitem{Zhang_IJCV21_ADRNet}
Pengyu Zhang, Dong Wang, Huchuan Lu, and Xiaoyun Yang.
\newblock Learning adaptive attribute-driven representation for real-time
  {RGB-T} tracking.
\newblock {\em International Journal of Computer Vision}, 129:2714–2729,
  2021.

\bibitem{Zhang_TIP21_JMMAC}
Pengyu Zhang, Jie Zhao, Dong Wang, Huchuan Lu, and Xiaoyun Yang.
\newblock Jointly modeling motion and appearance cues for robust {RGB-T}
  tracking.
\newblock {\em IEEE Transactions on Image Processing}, 30:3335 -- 3347, 2021.

\bibitem{Zhang_TCSVT_SiamCDA}
Tianlu Zhang, Xueru Liu, Qiang Zhang, and Jungong Han.
\newblock {SiamCDA}: Complementarity- and distractor-aware {RGB-T} tracking
  based on siamese network.
\newblock {\em IEEE Transactions on Circuits and Systems for Video Technology},
  99:1--16, 2021.

\bibitem{Zhang_CISP18_MDNet-RGBT}
Xingming Zhang, Xuehan Zhang, Xuedan Du, Xiangming Zhou, and Jun Yin.
\newblock Learning multi-domain convolutional network for {RGB-T} visual
  tracking.
\newblock In {\em International Congress on Image and Signal Processing,
  BioMedical Engineering and Informatics}, pages 1--6, 2018.

\bibitem{Zhang_ECCV20_Ocean}
Zhipeng Zhang, Houwen Peng, Jianlong Fu, Bing Li, and Weiming Hu.
\newblock Ocean: Object-aware anchor-free tracking.
\newblock In {\em European Conference on Computer Vision}, pages 771--787,
  2020.

\bibitem{zhu2020visDrone}
Pengfei Zhu, Longyin Wen, Dawei Du, Xiao Bian, Qinghua Hu, and Haibin Ling.
\newblock Vision meets drones: Past, present and future.
\newblock {\em arXiv preprint arXiv:2001.06303}, 2020.

\bibitem{Zhu_arxiv18_FANet}
Yabin Zhu, Chenglong Li, Yijuan Lu, Liang Lin, Bin Luo, and Jin Tang.
\newblock {FAN}et: Quality-aware feature aggregation network for {RGB-T}
  tracking.
\newblock {\em arXiv preprint arXiv:1811.09855}, 2018.

\bibitem{Zhu_MM19_DAPNet}
Yabin Zhu, Chenglong Li, Bin Luo, Jin Tang, and Xiao Wang.
\newblock Dense feature aggregation and pruning for {RGBT} tracking.
\newblock In {\em ACM International Conference on Multimedia}, pages 465--472,
  2019.

\end{thebibliography}


\begin{thebibliography}{1}\itemsep=-1pt

\bibitem{Kristan_ICCVW19_VOT19}
Matej Kristan, Jiri Matas, Ales Leonardis, Michael Felsberg, and et al.
\newblock The seventh visual object tracking {VOT2019} challenge results.
\newblock In {\em IEEE International Conference on Computer Vision Workshop},
  pages 1--36, 2019.

\bibitem{lukezic_CVPR2020_d3s}
Alan Lukezic, Jiri Matas, and Matej Kristan.
\newblock {D3S-A} discriminative single shot segmentation tracker.
\newblock In {\em IEEE Conference on Computer Vision and Pattern Recognition},
  pages 7133--7142, 2020.

\bibitem{wang_CVPR2019_siamMask}
Qiang Wang, Li Zhang, Luca Bertinetto, Weiming Hu, and Philip~HS Torr.
\newblock Fast online object tracking and segmentation: A unifying approach.
\newblock In {\em IEEE Conference on Computer Vision and Pattern Recognition},
  pages 1328--1338, 2019.

\bibitem{wang2019ranet}
Ziqin Wang, Jun Xu, Li Liu, Fan Zhu, and Ling Shao.
\newblock Ranet: Ranking attention network for fast video object segmentation.
\newblock In {\em IEEE International Conference on Computer Vision}, pages
  3978--3987, 2019.

\bibitem{yan_CVPR2021_alpha}
Bin Yan, Xinyu Zhang, Dong Wang, Huchuan Lu, and Xiaoyun Yang.
\newblock Alpha-refine: Boosting tracking performance by precise bounding box
  estimation.
\newblock In {\em IEEE Conference on Computer Vision and Pattern Recognition},
  pages 5289--5298, 2021.

\bibitem{Zhang_CVPR20_TVOS}
Yizhuo Zhang, Zhirong Wu, Houwen Peng, and Stephen Lin.
\newblock A transductive approach for video object segmentation.
\newblock In {\em IEEE Conference on Computer Vision and Pattern Recognition},
  pages 6949--6958, 2020.

\end{thebibliography}
}
\end{document}


\title{-- Supplementary Material -- \\
Visible-Thermal UAV Tracking: A Large-Scale Benchmark and New Baseline}

\maketitle

\section{Attribute Description of VTUAV}
We summarize the attribute description of VTUAV, which is shown in Tab.~\ref{Attribute description}. We state that comprehensive attribute annotation can fully exploit the effectiveness of attribute-aware trackers.
    \begin{table}[h]
		\caption{Attribute description in VTUAV.}\label{Attribute description}
		\vspace{-5mm}
		\footnotesize
		\begin{center}
			\begin{tabular}{cl}
				\toprule
				Name & Attribute description\\
				\midrule
				\textbf{TB} & \emph{Target Blur} - Target is blurry caused by illumination or motion.\\
				\textbf{CM} & \emph{Camera Movement} - Obvious camera motion occurs.\\
				\textbf{EI} & \emph{Extreme Illumination} - The target is in low or high light condition.\\
				\textbf{DEF} & \emph{Deformation} - The target is in no-rigid movement\\
				\textbf{PO} & \makecell[l]{\emph{Partial Occlusion} - The target is partial occluded by surroundings\\or some portion of target leaves the view.}\\
				\textbf{FO} & \emph{Full Occlusion} - The target is fully occluded by surroundings\\
				\textbf{SV} & \makecell[l]{\emph{Scale Variation} - The ratio of current bounding box and initial\\bounding box is out of range $\tau \in [0.5,2]$} \\
				\textbf{TC} & \makecell[l]{\emph{Thermal cluttering} - The background near the target has the\\similar temperature as the target.}\\
				\textbf{FM} & \makecell[l]{\emph{Fast Moving} - The offset of adjacent frames is larger than\\20 pixels.}\\
				\textbf{BC} & \makecell[l]{\emph{Background Clustering} - The background near the target has the\\similar color or texture as the target.}\\
				\textbf{OV} & \emph{Out-of-View} - The target is completely missing in current view.\\
				\textbf{LR} & \emph{Low Resolution} - The area of bounding box is less than 400. \\
				\textbf{TVS} &\makecell[l]{\emph{Thermal-Visible Separation} - The bounding boxes in visible and\\thermal images have no overlap.}\\
				\bottomrule
			\end{tabular}
		\end{center}
	\end{table}
	
\section{Attribute-based Comparison on VTUAV}
\begin{figure}[h]
    \centering
    \includegraphics[width=0.48\textwidth]{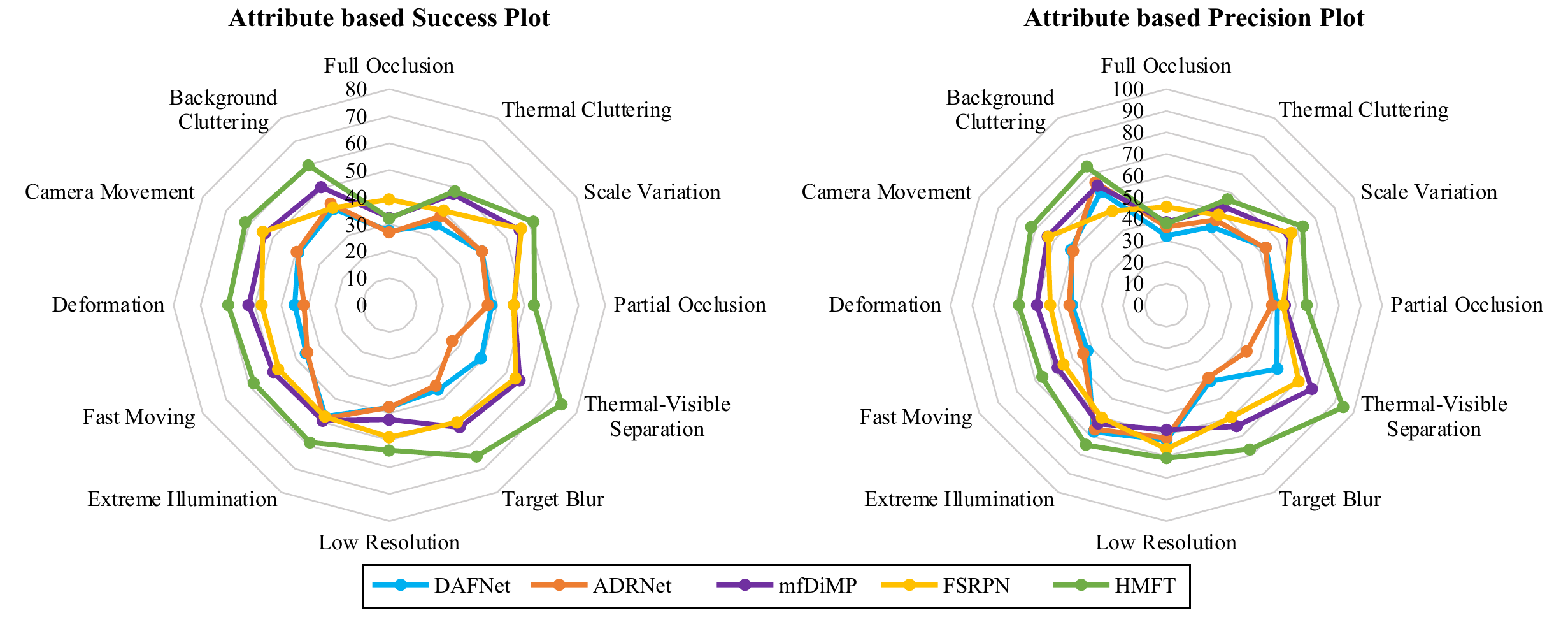}
    \caption{Attribute-based comparison on the short-term subset.}
    \vspace{-1.0em}
    \label{fig:attr_ST}
\end{figure}

\begin{figure}[h]
    \centering
    \includegraphics[width=0.48\textwidth]{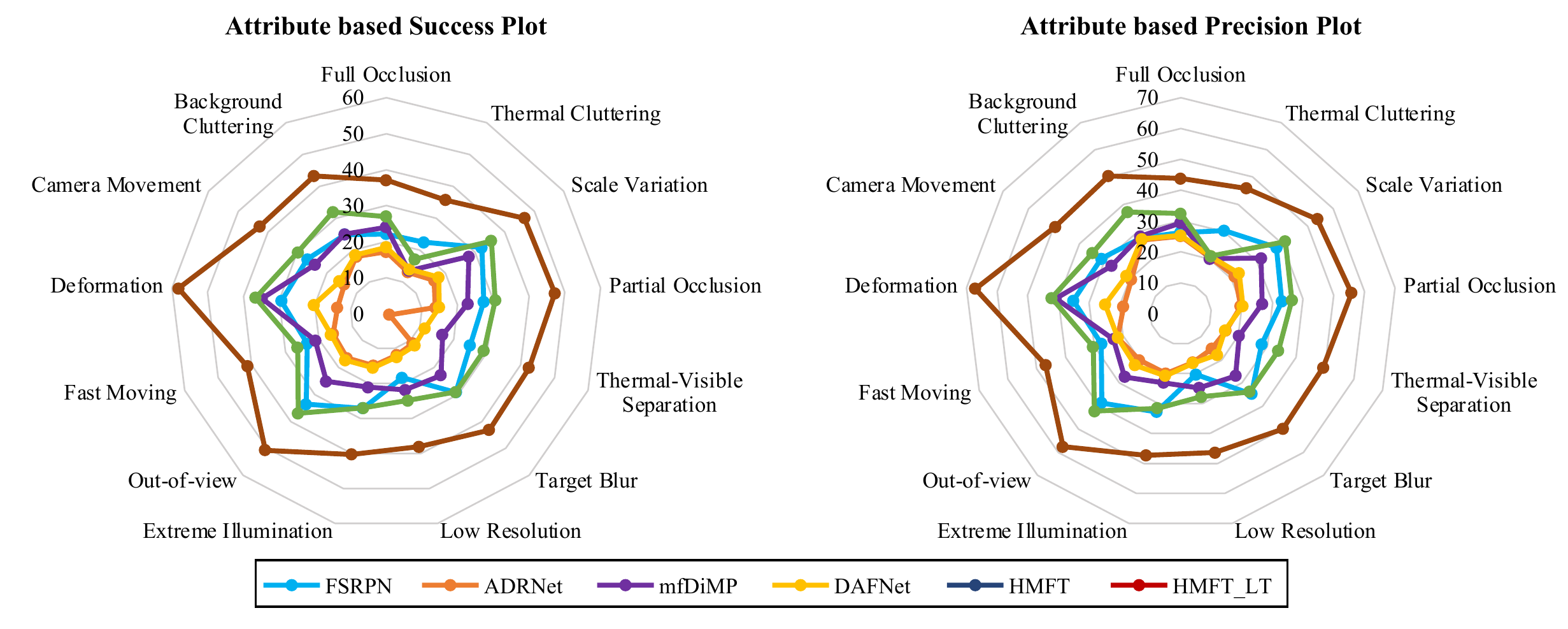}
    \caption{Attribute-based comparison on the long-term subset.}
    \vspace{-1.0em}
    \label{fig:attr_LT}
\end{figure}

We conduct attribute-specific analysis for both short-term and long-term trackers. As shown in Fig.~\ref{fig:attr_ST}, HMFT obtains the satisfying performance on all attributes, except fully occlusion, which is inferior to FSRPN~\cite{Kristan_ICCVW19_VOT19}. This phenomenon may stem from HMFT being an online tracker, where the absence of target involves the learning noise of filter. While FSRPN, which is a Siamese-based offline method, relies on the initial template, achieving a better re-detection ability. As shown in Fig.~\ref{fig:attr_LT}, the proposed long-term tracker HMFT\_LT shows its effectiveness on most of challenges, which outperforms the short-term version HMFT, with a large margin. This indicates that global mechanism achieves more than 10\% promotion in both MSR and MPR when the target is in out-of-view or fully occlusion. All the short-term trackers perform poor performance on TVS, TC and OV, which indicates the necessity of global detection component.

\section{Mask Evaluation}
We compare three short-term trackers~(SiamMask\cite{wang_CVPR2019_siamMask}, D3S~\cite{lukezic_CVPR2020_d3s} and AlphaRefine~\cite{yan_CVPR2021_alpha}) and two VOS methods~(TVOS~\cite{Zhang_CVPR20_TVOS} and RANet~\cite{wang2019ranet}), shown as Tab.~\ref{ours_sota_VOS}. All short-term trackers can output a mask for target with the initialization of bounding box. For VOS methods, we initialize the algorithms with the initial mask. VOS methods show inferior performance to short-term trackers with segmentation mask prediction. The reason is that short-term trackers predict the mask in a small search region, which avoid distractors with similar semantics, achieving better results.

\begin{table}[t]
	\caption{Mask evaluation results on the proposed dataset.}
	\vspace{-1.0em}
	\label{ours_sota_VOS}
	\small
	\begin{center}
		\begin{tabular}{lccc}
			\hline
			Method &  $\cal J$ & $\cal F$ & FPS\\
			\hline
			RANet~\cite{wang2019ranet} & 0.322 & 0.454 & 5.0\\
			TVOS~\cite{Zhang_CVPR20_TVOS} & 0.369 & 0.510 & 21.7\\
			SiamMask~\cite{wang_CVPR2019_siamMask} & 0.529 & 0.610 & \bf 57.2\\
			D3S~\cite{lukezic_CVPR2020_d3s} & 0.534 & 0.607 & 18.4\\
			AlphaRefine~\cite{yan_CVPR2021_alpha} & \bf 0.599 & \bf 0.719 & 14.2\\
			\hline
		\end{tabular}
		\vspace{-1.0em}
	\end{center}
\end{table}




\section{Analysis on Different Alignment Methods}
We apply image alignment for the initial frame of all the sequences. we compare the center distance among previous datasets~(GTOT and RGBT234), shown in Tab.~\ref{tab:center_dist}. We show that our alignment method is also effective for image registration. In the future, trackers can design an alignment module for better results. 
\begin{table}[h]
	\caption{Comparison among previous datasets on center distance of bounding boxes in visual-thermal modalities.}
	\label{tab:center_dist}
	\small
	\begin{center}
		\begin{tabular}{ccccccc}
			\toprule
			&\multicolumn{2}{c}{GTOT} &
			\multicolumn{2}{c}{RGBT234} & \multicolumn{2}{c}{VTUAV}\\
			&Mean & Median & Mean & Medsian & Mean & Median \\
			\midrule
			Dist. & 2.74 & 2.81 & 5.19 & 4.54 & 10.99 & 8.83\\
			\bottomrule
		\end{tabular}
	\end{center}
\end{table}
\section{Visualization Results on Long-term Subset}
\begin{figure}[t]
    \caption{Visualization of HMFT and HMFT\_LT on long-term subset of VTUAV.}
    \label{QA_LT}
    \includegraphics[width=0.45\textwidth]{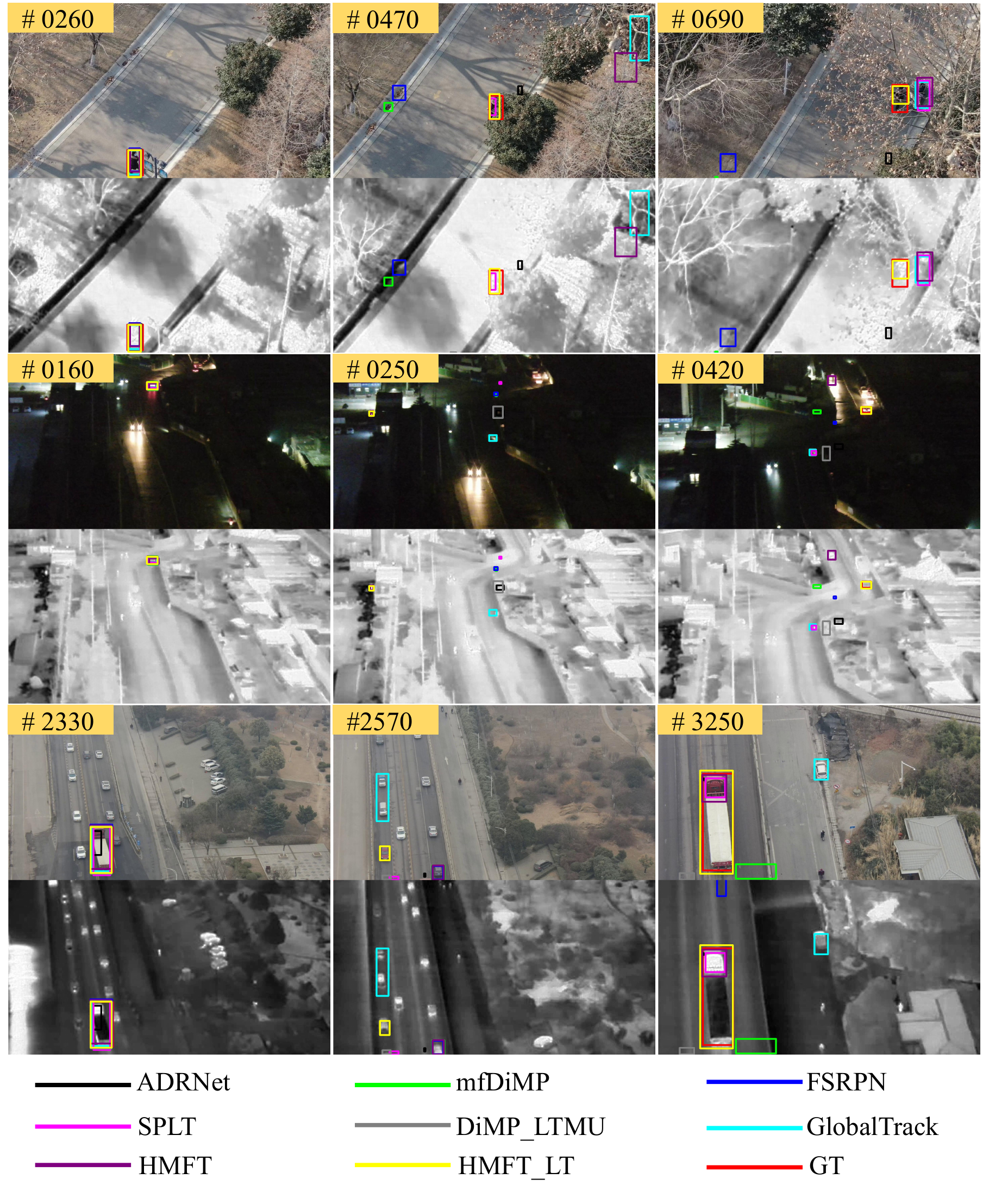}
\end{figure}

As shown in Fig.~\ref{QA_LT}, we reveal the effectiveness of our HMFT trackers on long-term subset of VTUAV. Compared with existing long-term trackers, our tracker can handle all the challenging cases in various conditions, and show its great potential on re-detection ability. 

\section{Failure Cases}
\begin{figure}[t]
    \centering
    \includegraphics[width=0.45\textwidth]{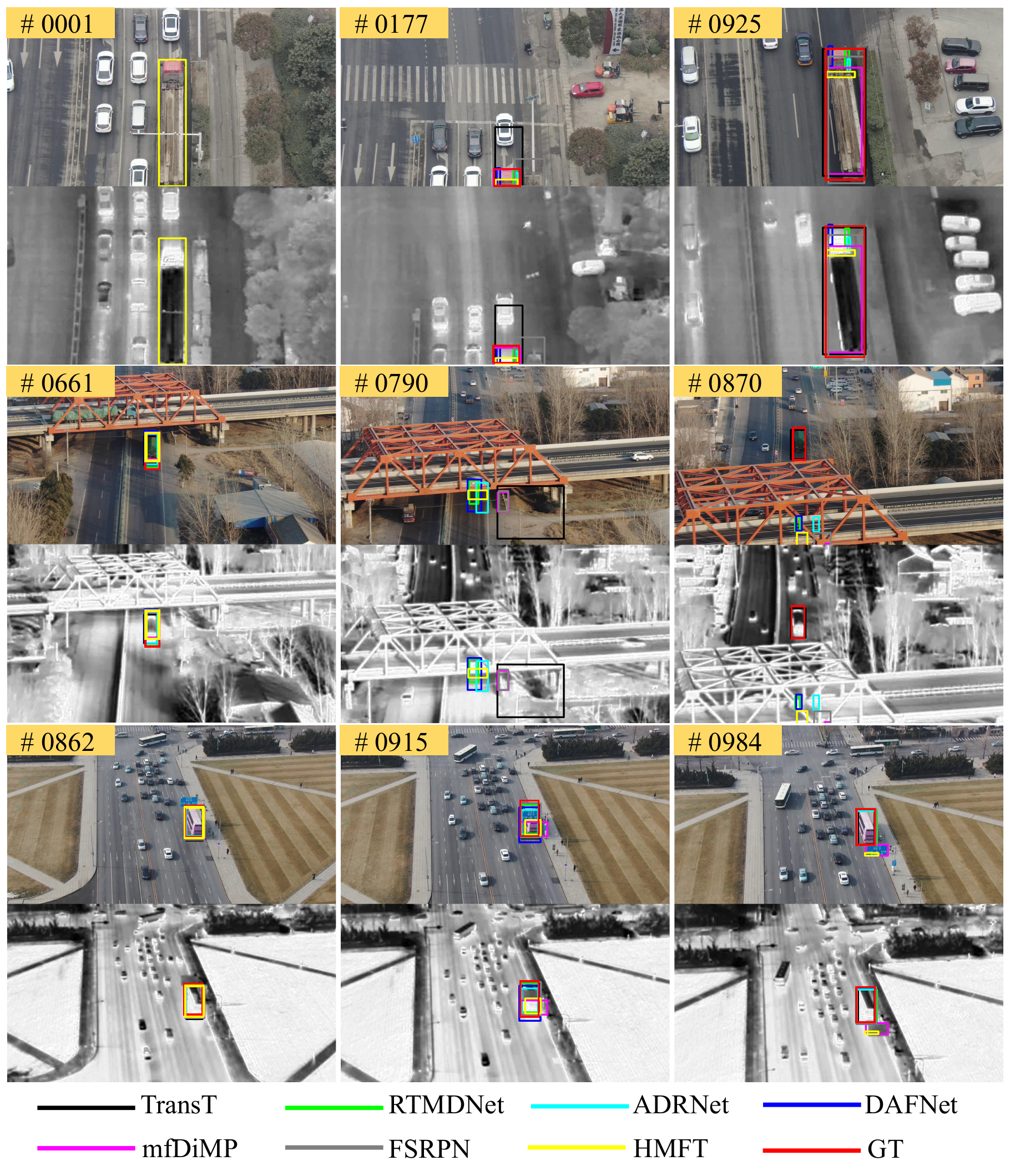}
    \caption{Failure analysis of the proposed HMFT.}
    \vspace{-1.0em}
    \label{failure_case}
\end{figure}

As shown in Fig.~\ref{failure_case}, we provide the limitation of HMFT tracker, HMFT often suffers target missing when the target is in heavy occlusion or full occlusion, leading to an inferior performance on fully occlusion attributes~(which is shown in Figure~4 in the submission). When the target is occluded, our tracker has limited ability on re-detection. The reason is that the noisy samples cause biased filter learning in the online learning process, while Siamese-based methods in an offline manner get rid of this issue. Our long-term verision~(HMFT\_LT) with a global tracker further improves our method in this challenge. In the future, development, such as the adaptive updating mechanism, can be accomplished to achieve better performance.

{\small
\bibliographystyle{ieee_fullname}
\bibliography{egbib}
}